\pdfoutput=1
\documentclass[11pt, a4paper, copyright, gdm]{google}

\usepackage[authoryear, round]{natbib}
\expandafter\def\expandafter\UrlBreaks\expandafter{\UrlBreaks\do-}
\usepackage{booktabs}
\usepackage{graphicx}
\usepackage{float}
\usepackage{listings}
\usepackage{tabularx}
\usepackage{array}
\usepackage{multirow}
\newcommand{\thead}[1]{\begin{tabular}[c]{@{}c@{}}#1\end{tabular}}
\usepackage{xcolor}

\makeatletter\g@addto@macro\bfseries{\boldmath}\makeatother
\newcommand{\tablenotes}[1]{\par\vspace{4pt}\parbox{\textwidth}{\footnotesize #1}}
\newcommand{\hfecaret}{\kern-0.13em\textasciicircum\kern-0.15em}
\makeatletter\@clubpenalty=\clubpenalty\makeatother
\setlist{beginpenalty=10000}
\let\hfeoldsection\section
\renewcommand{\section}{\Needspace{7\baselineskip}\hfeoldsection}
\let\hfeoldsubsection\subsection
\renewcommand{\subsection}{\Needspace{6\baselineskip}\hfeoldsubsection}

\uselogo{}

\title{Humanity's First Exam: Measuring the Reliability Floor of LLMs}
\hypersetup{pdftitle={Humanity's First Exam: Measuring the Reliability Floor of LLMs},
  pdfauthor={Oran Kelly, Anastasios Kementsietsidis, Chrysovalantis Anastasiou, Alex Yakubovich, Beat Weichsler, Natalie Ha, Ryan Burnell, Samuel Albanie, Orhan Firat}}

\correspondingauthor{orankelly@google.com}

\renewcommand{\today}{2026-11-24}

\author[1]{Oran~Kelly}
\author[1]{Anastasios~Kementsietsidis}
\author[1]{Chrysovalantis~Anastasiou}
\author[2]{Alex~Yakubovich}
\author[3]{Beat~Weichsler}
\author[2]{Natalie~Ha}
\author[1]{Ryan~Burnell}
\author[1]{Samuel~Albanie}
\author[1]{Orhan~Firat}

\affil[1]{Google DeepMind}
\affil[2]{Google Cloud}
\affil[3]{Google Research}

\begin{abstract}
Frontier Large Language Models (LLMs) now achieve breakthrough-level performance across domains such as mathematics, programming, and science. However, these same models can still occasionally fail on elementary tasks that humans can easily solve. We argue this is a byproduct of modern evaluation design. Standard benchmarks typically measure the ``ceiling'' capability of models via single-shot (pass@1) or multi-sample (pass@k) accuracy on increasingly specialized tasks, hiding deficits in floor reliability.\par\vspace{4pt}\noindent
To address this gap, we introduce Humanity's First Exam (HFE), a benchmark consisting of 544 text and multimodal prompts across 20 foundational subcategories. Prompts are organized into four-prompt families, each built from a seed prompt filtered through three overlapping criteria (Simplicity, Relatability, and Shock Value) so that they are easy to explain, straightforward to solve, and surprising when a frontier model fails. Evaluating 18 frontier and fast-tier models from leading providers, we find that high single-shot accuracy masks substantial reliability deficits when models must succeed consistently across repeated rollouts and closely related variants. The top five models reach 96\% to 98\% pass@1 and 100\% family pass@4, yet average only 85.8\% on family pass\hfecaret{}4 (80.5\% to 92.4\%), with a 7.6 to 80.1 percentage-point Family Reliability Gap across all 18 models. Remaining failures are heavily concentrated in visual perception (73\% of wrong answers), where extra test-time reasoning yields far smaller gains than on text and symbolic tasks. Real-world deployment requires evaluations that enable us to measure and raise the reliability floor, not just the capability ceiling. Therefore, we release the HFE \href{https://www.kaggle.com/benchmarks/deepmind/hfe/leaderboard}{benchmark leaderboard} on Kaggle, alongside a stratified $\sim$20\% \href{https://www.kaggle.com/datasets/deepmind/hfe-public}{public subset} of evaluation prompts.\par\vspace{3pt}\noindent
\end{abstract}

\makeatletter
\g@addto@macro\ps@firststyle{%
  \fancyfoot[L,C,R]{}%
}
\newcommand{\firstpagefooterblock}{%
  \begingroup
  \NoHyper
  \global\let\orig@footnoterule\footnoterule
  \gdef\footnoterule{\kern-3pt{\color{gray}\hrule width \textwidth height 1pt}\kern 4pt\global\let\footnoterule\orig@footnoterule}%
  \long\def\@makefntext##1{\parindent 0pt\parskip 2pt\noindent\footerfont##1}%
  \footnotetext{%
    {\itshape Corresponding author(s): \the\correspondingauthor}%
    \ifthenelse{\boolean{copyright}}{\\\copyrightext}{}%
  }%
  \endNoHyper
  \endgroup
}
\makeatother

\begin{document}
\maketitle
\enlargethispage{24pt}
\firstpagefooterblock

\vspace{-2pt}
\begin{figure}[H]
  \centering
  \captionsetup{font=small}
  \includegraphics[width=0.80\textwidth]{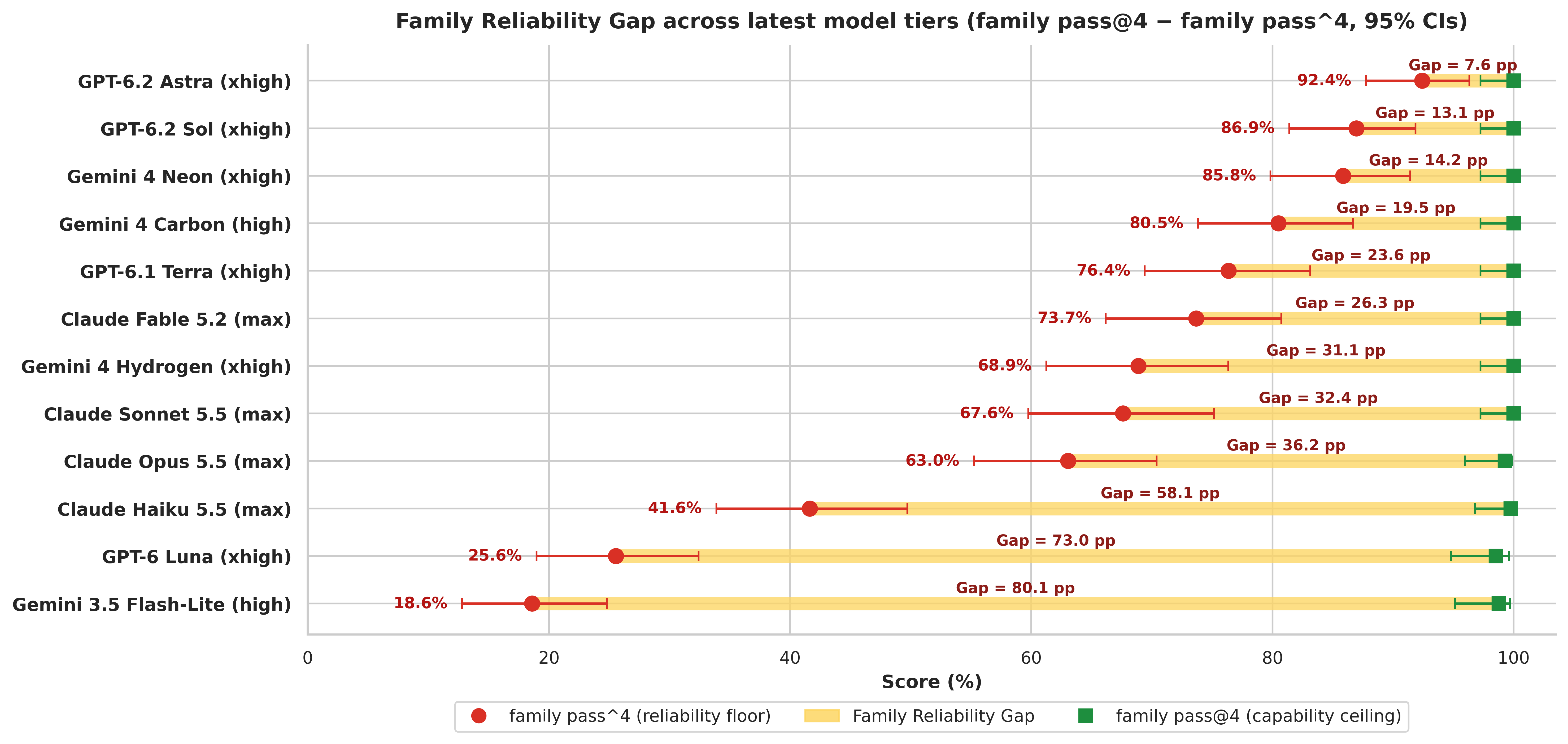}
  \vspace{-5pt}
  \caption{Family Reliability Gap (family pass@4 $-$ family pass\hfecaret{}4, with 95\% CIs) across 12 representative frontier and fast-tier models from Google, OpenAI, and Anthropic (four per provider). While models nearly saturate the capability ceiling (family pass@4 = 98.5\% to 100\%), floor reliability drops sharply (family pass\hfecaret{}4 = 18.6\% to 92.4\%), leaving a 7.6 to 80.1~pp gap. Complete results appear in Table~\ref{tab:main_results} and Appendix Figure~\ref{fig:reliability_gap_full18}.}
  \label{fig:hero_gap}
\end{figure}
\clearpage

\section{Introduction}
\label{sec:intro}

Recent LLM progress has been defined by rapid gains at the boundaries of model capabilities, from gold-medal performance at the ICPC World Finals~\citep{deepmind2025icpc} to AI-assisted, formally verified progress on the Navier--Stokes problem~\citep{openai2026a_navierstokes} and a reported Lean~4 formalization of the Generalized Riemann Hypothesis~\citep{alpoge2026remark}. Yet these same models can still fail at tasks people consider simple or routine. This discrepancy reflects the ``Jagged Frontier'' of artificial intelligence~\citep{dellacqua2023jagged,karpathy2024jagged}, the counterintuitive phenomenon whereby a model's capability boundary is non-uniform, encompassing tasks of high technical complexity while excluding seemingly trivial and cognitively simple ones.

\begin{figure}[!htbp]
  \centering
  \begin{tabular*}{\textwidth}{@{}>{\raggedright\arraybackslash}p{0.482\textwidth}@{\extracolsep{\fill}}>{\raggedright\arraybackslash}p{0.482\textwidth}@{}}
    \includegraphics[width=\linewidth]{figures/fig2a_mugs.png} &
    \includegraphics[width=\linewidth]{figures/fig2b_neon.png} \\[4pt]
    \footnotesize Prompt 1: ``How many coffee mugs in the forefront do not have steam rising from them?'' &
    \footnotesize Prompt 2: ``List the color of the third letter of each word in the neon sign in order, separated by a semicolon.''
  \end{tabular*}
  \caption{Two example prompts from HFE's public split, each drawn from a four-prompt task family: Object Counting (left; answer: 4) and Color Identification (right; answer: green; yellow; white; blue; blue). Both are straightforward for humans, yet multiple models across providers fail them; the complete Color Identification family is shown in the Appendix.}
  \label{fig:examples}
\end{figure}

LLM benchmarks currently tend to focus on model capability ceilings and efficiency metrics, but another critical component of model performance is baseline reliability. \emph{Inconsistent} model performance (i.e., succeeding on one or even many attempts at a task, but not every time, or solving one task conclusively while failing a very closely related one) is likely to cause confusion and frustration about where LLMs are worth deploying. Moreover, for many automated workflows, time-sensitive tasks, and tasks of critical importance, LLMs can only be incorporated if people have high confidence the model will reliably complete the task without failures. Despite the importance of reliability, few frontier benchmarks explicitly measure multi-run consistency~\citep{yao2024taubench,barres2025tau2bench,allcock2026obviousbench} or variant invariance~\citep{mirzadeh2024gsmsymbolic,srivastava2024functional,luo2026jaggedbench}, and to our knowledge none evaluate both jointly across foundational text and multimodal tasks.

To address this blind spot, we introduce Humanity's First Exam (HFE), a benchmark designed to measure the reliability floor of frontier models (Figure~\ref{fig:hero_gap}; example prompts in Figure~\ref{fig:examples}). Rather than assessing whether a model can solve a difficult task at least once after multiple retries (pass@k), HFE evaluates whether a model can \emph{consistently} solve an elementary task across both repeated runs and closely related variants (pass\hfecaret{}k and family pass\hfecaret{}k; building on \citealp{yao2024taubench}). Throughout this paper, reliability refers to whether a model consistently produces correct outputs, rather than serving or infrastructure stability. While reliability is equally critical in agentic workflows (\S\ref{sec:related-agentic}), we scope HFE to single-turn tasks so that results reflect the model rather than external tools or agent harnesses.

In this paper, alongside introducing the HFE dataset and evaluation methodology, we investigate three core empirical questions:
\begin{enumerate}[leftmargin=*, itemsep=2pt]
  \item \textbf{How low is the reliability floor across frontier LLMs?} To what extent do today's leading models fail on elementary tasks where perfect consistency is expected, and how does that floor differ across smaller and larger model tiers?
  \item \textbf{How wide is the reliability gap between the capability ceiling (pass@4, family pass@4) and reliability floor (pass\hfecaret{}4, family pass\hfecaret{}4)?} Do models fail elementary tasks because they lack the underlying capability, or because they fail to leverage this capability consistently?
  \item \textbf{Where do reliability gaps concentrate, and how much does test-time reasoning help?} Which task categories and modalities drive floor failures, how do model errors compare with human difficulty, and to what extent does increased reasoning effort close the gap?
\end{enumerate}

\section{Benchmark Design}
\label{sec:design}

Traditional LLM benchmark design prioritizes discriminative power at the upper boundary of model performance. HFE inverts this approach, measuring the baseline reliability of frontier models on elementary tasks that a majority of humans can complete.

\subsection{Taxonomy}
\label{sec:taxonomy}

To ground our taxonomy in real-world failures, we began by curating widely shared social-media posts (X, Reddit) following frontier model releases that showed elementary LLM errors across five broad areas: logical reasoning, visual perception, basic arithmetic and counting, constraint adherence, and false-premise questions (excluding image generation tasks to focus on objectively gradable text outputs). Expanding these areas into an exploratory 500-prompt pilot set (drawn from social media, public failure lists, and elementary-level reasoning exams) revealed that coarse categories frequently entangled multiple skills within a single prompt. Fine-grained capability annotation and clustering of this 500-prompt pilot set yielded 23 atomic, single-skill sub-abilities (Table~\ref{tab:categories}).

\begin{table}[!htbp]
  \centering
  \caption{Categories of tasks in HFE.}
  \label{tab:categories}
  \scriptsize
  \setlength{\tabcolsep}{4pt}
  \renewcommand{\arraystretch}{1.12}
  \begin{tabularx}{\textwidth}{@{}r >{\raggedright\arraybackslash}p{0.27\textwidth} >{\raggedright\arraybackslash}X l l@{}}
    \toprule
    \textbf{\#} & \textbf{Category Name (Families / Prompts)} & \textbf{Description} & \textbf{Input Modality} & \textbf{Macro-Category} \\
    \midrule
    1 & \textbf{Algorithmic and Symbolic Execution (8 / 32)} & Manipulating symbols strictly by rule, ignoring semantic meaning. & Text & Math \& Algorithmic \\
    2 & \textbf{Analogical Reasoning (Structural)\textsuperscript{\textdagger} (0 / 0)} & Identifying abstract relationships (A is to B as C is to D). & Text, Text \& Image & Reasoning \& Language \\
    3 & \textbf{Causal and Counterfactual Reasoning (2 / 8)} & Understanding cause-and-effect and hypotheticals. & Text, Text \& Image & Reasoning \& Language \\
    4 & \textbf{Commonsense and Physical Reasoning (4 / 16)} & Basic physics, gravity, and object permanence. & Text, Text \& Image & Reasoning \& Language \\
    5 & \textbf{Constraint Satisfaction (13 / 52)} & Following strict negative or formatting instructions that require suppressing the model's natural output. & Text & Math \& Algorithmic \\
    6 & \textbf{General Arithmetic and Calculation (3 / 12)} & Basic integer math, money, and simple addition. & Text & Math \& Algorithmic \\
    7 & \textbf{Generalization and Pattern Recognition (4 / 16)} & Identifying a rule from small examples and applying it to a new instance. & Text, Text \& Image & Reasoning \& Language \\
    8 & \textbf{Inductive and Probabilistic Reasoning\textsuperscript{\textdagger} (0 / 0)} & Determining the most likely outcome based on evidence; distinguishing probability from certainty. & Text & Reasoning \& Language \\
    9 & \textbf{Logical Deduction (2 / 8)} & Multi-step transitive logic. & Text & Reasoning \& Language \\
    10 & \textbf{Mathematical and Quantitative Reasoning (5 / 20)} & Understanding math concepts beyond simple calculation (magnitude, comparison, units). & Text, Text \& Image & Math \& Algorithmic \\
    11 & \textbf{Positional Tracking (6 / 24)} & Keeping track of object state/location through a sequence of moves. & Text, Text \& Image & Spatial \& Positional \\
    12 & \textbf{Sequence Manipulation (6 / 24)} & Reordering, reversing, or editing strings. & Text & Math \& Algorithmic \\
    13 & \textbf{Sorting and Ordering (5 / 20)} & Organizing lists by criteria (alphabetical, size). & Text, Text \& Image & Math \& Algorithmic \\
    14 & \textbf{Spatial and Topological Reasoning (4 / 16)} & Understanding geometry, shapes, and how objects fit together. & Text, Text \& Image & Spatial \& Positional \\
    15 & \textbf{Temporal Arithmetic\textsuperscript{\textdagger} (0 / 0)} & Calculations involving dates, times, and calendars. & Text, Text \& Image & Math \& Algorithmic \\
    16 & \textbf{Negative Fact Verification (3 / 12)} & Correctly identifying when something did not happen or does not exist. & Text, Text \& Image & Reasoning \& Language \\
    17 & \textbf{Pragmatic Reasoning (3 / 12)} & Grasping implied meaning, obvious sarcasm, or the practical intent behind a statement rather than taking it literally. & Text & Reasoning \& Language \\
    18 & \textbf{Color Identification (10 / 40)} & Correctly identifying specific colors/shades in an image, resisting lighting artifacts or common priors. & Text \& Image & Vision \\
    19 & \textbf{Object Counting (12 / 48)} & Accurately counting distinct objects. & Text \& Image & Vision \\
    20 & \textbf{Text \& Symbol Recognition (11 / 44)} & Reading text (OCR) in images, especially handwriting or distorted text. & Text \& Image & Vision \\
    21 & \textbf{Arrows, pointers, or indicators (11 / 44)} & Interpreting the meaning or direction of superimposed graphical elements (drawn arrows, circles) or physical pointing. & Text \& Image & Vision \\
    22 & \textbf{Graph \& Chart Understanding (7 / 28)} & Reading basic visual data points, legends, or obvious visual trends from simple charts. & Text \& Image & Vision \\
    23 & \textbf{Spatial Reasoning (17 / 68)} & Understanding 3D physical relationships, occlusion, or relative positioning of objects depicted in a 2D image or described scene (e.g., inside/outside, in front of/behind, stacking). & Text, Text \& Image & Spatial \& Positional \\
    \bottomrule
  \end{tabularx}
  \tablenotes{\textsuperscript{\textdagger}~Three pilot subcategories (Analogical Reasoning (Structural), Inductive and Probabilistic Reasoning, and Temporal Arithmetic), included during seed authoring (\S\ref{sec:seeds}) and human baseline collection (\S\ref{sec:human-baseline}), had no surviving families after quality subsampling (\S\ref{sec:subsampling}), leaving 20 subcategories in the final benchmark.}
\end{table}

\subsection{Seed Prompts}
\label{sec:seeds}

We authored new, unleaked seed prompts for each subcategory. Candidate seeds had to satisfy three requirements:
\begin{enumerate}[leftmargin=*, itemsep=2pt]
  \item \textbf{Single sub-ability:} Targets one isolated sub-ability so failures are diagnostic.
  \item \textbf{Empirical failure on calibration panel:} Elicits at least one error across four sampled rollouts on at least one model in our seed calibration panel (\textbf{Gemini~3.1 Flash-Lite}, \textbf{Gemini~3.1 Pro}, \textbf{Claude Opus~4.7}, \textbf{Claude Sonnet~4.6}, and \textbf{GPT-5.5}), filtering out already-saturated tasks at creation time (May to June 2026).
  \item \textbf{Virality criteria:} Passes all three binary review tags; \textbf{(a) Simplicity:} The task and failure mode can be explained easily; \textbf{(b) Relatability:} The task is easily human-solvable without domain expertise, using common sense or perception; and \textbf{(c) Shock Value:} A frontier model failure appears logically absurd and starkly incompatible with its perceived capability ceiling.
\end{enumerate}
This process produced \textbf{321 seed prompts} satisfying all criteria across the 23 subcategories (Table~\ref{tab:categories}).

\subsection{Prompt Families}
\label{sec:families}

Evaluating models on isolated phrasings risks conflating task competence with lucky surface wording. To test structural invariance, each seed prompt is expanded into a five-prompt \textbf{Prompt Family} consisting of \textbf{one seed prompt} and \textbf{four variants}: controlled surface edits (renaming entities, reordering clauses, or varying visual layout/attributes) that preserve the underlying sub-ability, cognitive difficulty, and task logic (not necessarily answer preserving).

During variant authoring (June 26 to July 26, 2026), candidate variants were calibrated across four rollouts per model on the initial five-model panel augmented with two newly released models (\textbf{Gemini~3.5 Flash} and \textbf{Claude Opus~4.8}); a variant was retained only if it elicited at least one failure on this seven-model panel. \textbf{Six} of the 321 seeds for which four model-failing variants could not be constructed were discarded during family generation, yielding \textbf{315 complete five-prompt families (1,575 prompts)}. \emph{(Of the 18 models evaluated in Table~\ref{tab:main_results}, 16 were released after data collection concluded, while Gemini~3.5 Flash-Lite and Gemini~3.6 Flash launched on July 21, 2026, in the final week of variant creation. Gemini~3.5 Flash was subsequently deprecated and routed to Gemini~3.6 Flash prior to our evaluation window and is therefore omitted from Table~\ref{tab:main_results}.)}

\subsection{Human Baseline}
\label{sec:human-baseline}

To verify that candidate tasks are easily solvable for humans without tools, matching model evaluation conditions, each seed prompt was solved independently by four humans, excluding anyone who had authored or previously reviewed the task. Raters worked self-paced using only pen and paper or a plain scratchpad, strictly prohibiting LLMs, web search, calculators, code, word or character counters, copy-paste, and writing assistants, with their completion time being logged. An independent human reviewer then graded all four human responses against the ground truth, tagging mismatches to separate genuine human errors from prompt or ground truth issues (e.g., ambiguity). Across all 1,260 human responses on these 315 seed prompts, single-rater accuracy was 80.3\% with a median per-seed solve time of 1.5~minutes (IQR 1.0 to 2.5): 79.7\% of seeds (251/315) were solved by at least 3 of 4 raters (50.5\% at 4/4 and 29.2\% at 3/4), whereas 20.3\% fell below 3/4 and were subsequently removed during quality subsampling (\S\ref{sec:subsampling}). On the 136 seeds retained in the final benchmark (\S\ref{sec:subsampling}), single-rater accuracy is 90.8\% (494/544, with 86 seeds at 4/4 and 50 at 3/4, and all 50 misses attributed to human rater error rather than prompt ambiguity); median per-rater solve time is 1.0~minute on both Vision and Spatial \& Positional seeds (IQR 1.0 to 2.0) and 1.0~minute across the 86 unanimous 4/4 seeds.

\subsection{Quality Subsampling}
\label{sec:subsampling}

Starting from the \textbf{315 seed prompts (1,575 total prompts including all variants)} satisfying all three virality criteria (\emph{Simplicity}, \emph{Relatability}, and \emph{Shock Value}), we enforce three sequential filtering steps (Table~\ref{tab:filtering}): (1)~\textbf{Human simplicity}, retaining only families whose seed prompt was solved by at least 3 of 4 independent human raters (\textbf{251 families / 1,255 prompts}); (2)~\textbf{QA audit}, auditing seeds, variants, autorater edge cases, and model-flagged items, removing the entire family whenever the seed prompt (or multiple variants) exhibited visual or textual ambiguity, an unfaithful variant shift, or an inaccurate ground truth (\textbf{$-$115 families, leaving 136 families / 680 prompts}); and (3)~\textbf{Variant standardization}, dropping one variant per remaining family (any single variant flagged during QA where present, and otherwise the highest-pass-rate redundant variant) so every family contains four verified prompts (\textbf{1 seed + 3 variants}). This yields a final benchmark of \textbf{136 complete four-prompt families (544 prompts)}.

\begin{table}[!htbp]
  \centering
  \caption{Dataset filtering process.}
  \label{tab:filtering}
  \footnotesize
  \setlength{\tabcolsep}{5pt}
  \renewcommand{\arraystretch}{1.15}
  \begin{tabularx}{\textwidth}{@{}c >{\raggedright\arraybackslash}X r r l@{}}
    \toprule
    \textbf{Step} & \textbf{Rule} & \textbf{Families} & \textbf{Prompts} & \textbf{Dropped} \\
    \midrule
    0 & Initial complete 5-prompt families (\S\ref{sec:families}) & 315 & 1,575 & --- \\
    1 & \textbf{Human simplicity:} At least 3 of 4 human raters solve the seed prompt & 251 & 1,255 & \begin{tabular}[t]{@{}l@{}}64 families\\320 prompts\end{tabular} \\
    2 & \textbf{QA audit:} Remove the entire family if the seed prompt (or $\ge$2 variants) had prompt/ground-truth ambiguity, autorater issues, or unfaithful variant shifts & 136 & 680 & \begin{tabular}[t]{@{}l@{}}115 families\\575 prompts\end{tabular} \\
    3 & \textbf{Variant standardization:} Drop 1 variant per remaining family (QA-flagged or highest-pass-rate redundant variant) to form uniform 4-prompt families & 136 & 544 & 136 prompts (1 per family) \\
    \bottomrule
  \end{tabularx}
\end{table}

\subsection{Final Dataset}
\label{sec:final-dataset}

The final 544-prompt HFE dataset spans 20 foundational subcategories; three pilot subcategories had no remaining families after quality subsampling (Table~\ref{tab:categories}). By input modality, HFE comprises 316 multimodal (Text \& Image) prompts across 79 families (58.1\%) and 228 text-only prompts across 57 families (41.9\%), with every family sharing a consistent modality across its four prompts. To support community reproducibility while mitigating benchmark contamination, we release a stratified $\sim$20\% public subset of 112 prompts (28 complete families) under CC-BY-4.0 on Kaggle (\url{https://www.kaggle.com/datasets/deepmind/hfe-public}) and maintain the remaining 432 prompts (108 families) as a private holdout (\S\ref{sec:release-subset}), with live scores hosted on the Kaggle HFE leaderboard (\url{https://www.kaggle.com/benchmarks/deepmind/hfe/leaderboard}).

\section{Evaluation Methodology}
\label{sec:methodology}

\subsection{Metrics}
\label{sec:metrics}

For each prompt, we sample $n = 6$ independent generations and let $c \in \{0, \dots, n\}$ denote the number of correct responses.

\paragraph{\textit{Prompt-level metrics.}}
Standard evaluations report $\mathrm{pass}@k$~\citep{chen2021codex}, the probability that \emph{at least one} of $k$ draws without replacement is correct ($\mathrm{pass}@1 = c/n$ measures single-shot accuracy; $\mathrm{pass}@4$ measures multi-sample capability ceiling). Because elementary tasks require reliable execution on every call, we follow \citet{yao2024taubench} and measure operational consistency via $\mathrm{pass}^{k}$, the probability that \emph{all} $k$ draws without replacement are correct:
\begin{equation*}
  \mathrm{pass}@k = 1 - \prod_{i=0}^{k-1} \frac{n - c - i}{n - i},
  \qquad
  \mathrm{pass}^{k} = \prod_{i=0}^{k-1} \frac{c - i}{n - i}
\end{equation*}
At $k = 1$, $\mathrm{pass}@1 = \mathrm{pass}^{1} = c/n$; as $k$ grows, $\mathrm{pass}@k$ increases (capability) while $\mathrm{pass}^{k}$ decreases (consistency). We report $\mathrm{pass}^{4}$ for single-prompt reliability and define the \emph{Single-Prompt Reliability Gap} as $\mathrm{pass}@k - \mathrm{pass}^{k}$.

\begin{figure}[!htbp]
  \centering
  \includegraphics[width=0.88\textwidth]{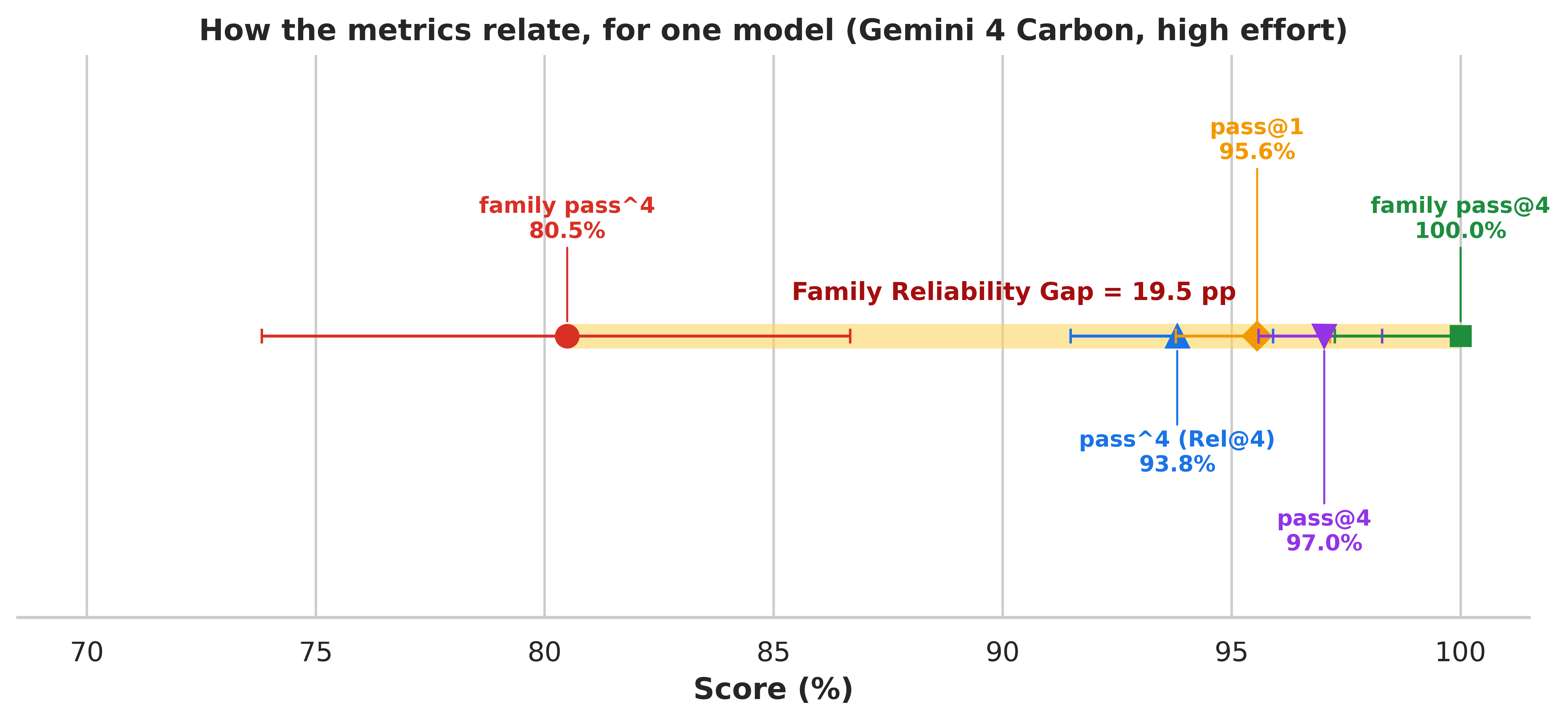}
  \caption{The five metrics for a single model (Gemini~4 Carbon, high effort), with 95\% CIs; the x-axis starts at 70\%. The metrics are ordered as family pass\hfecaret{}4 $\le$ pass\hfecaret{}4 $\le$ pass@1 $\le$ pass@4 $\le$ family pass@4 for every model; the shaded bar is the Family Reliability Gap.}
  \label{fig:metrics_single_model}
\end{figure}

\paragraph{\textit{Family-level metrics.}}
To measure robustness across minor prompt variations, each task family $f \in \{1, \dots, M\}$ ($M = 136$) comprises four logic-preserving prompts (one seed and three variants) $j \in \{1, \dots, 4\}$. Over a joint $4 \times k$ draw ($k$ rollouts per prompt), we define:
\begin{equation*}
  \mathrm{family\ pass}@k(f) = 1 - \prod_{j=1}^{4} \bigl(1 - \mathrm{pass}@k(f, j)\bigr),
  \qquad
  \mathrm{family\ pass}^{k}(f) = \prod_{j=1}^{4} \mathrm{pass}^{k}(f, j)
\end{equation*}
Here, $\mathrm{family\ pass}@1$ ($4 \times 1$) and $\mathrm{family\ pass}@4$ ($4 \times 4$) test whether \emph{at least one} rollout succeeds across the family, $\mathrm{family\ pass}^{1}$ ($4 \times 1$) isolates \emph{single-draw variant invariance} across all four prompts, and $\mathrm{family\ pass}^{4}$ ($4 \times 4$) measures \emph{joint rollout-and-variant reliability}. Prompt and family metrics are averaged over all 544 prompts and 136 families, respectively, with 95\% confidence intervals via family-clustered bootstrap. Hence we define the \emph{Family Reliability Gap} as $\mathrm{family\ pass}@k - \mathrm{family\ pass}^{k}$. Figure~\ref{fig:metrics_single_model} illustrates how these metrics relate for a single model, and Table~\ref{tab:metrics} summarizes all six definitions.

\begin{table}[!htbp]
  \centering
  \caption{Summary of HFE evaluation metrics. Each row defines a metric for a single unit (either a single prompt for rows 1 to 3, or a single family for rows 4 to 6). Reported scores are averaged over all 544 prompts (prompt-level metrics) or all 136 families (family-level metrics); 95\% CIs are from a family-clustered bootstrap.}
  \label{tab:metrics}
  \footnotesize
  \setlength{\tabcolsep}{5pt}
  \renewcommand{\arraystretch}{1.10}
  \begin{tabularx}{\textwidth}{@{}c l l >{\raggedright\arraybackslash}X@{}}
    \toprule
    & \textbf{Metric (Scope)} & \textbf{Formula ($n = 6$)} & \textbf{What It Measures} \\
    \midrule
    \textbf{1} & \textbf{pass@1} (Single prompt) & $c / n$ & \textbf{Single-try accuracy:} Probability \textbf{1 random rollout} is correct. Standard headline metric. \\
    \textbf{2} & \textbf{pass@4} (Single prompt) & $1 - C(n-c, 4) / C(n, 4)$ & \textbf{Capability ceiling:} Probability \textbf{at least 1 of 4 rollouts} is correct. \\
    \textbf{3} & \textbf{pass\hfecaret{}4 (Rel@4)} (Single prompt) & $C(c, 4) / C(n, 4)$ & \textbf{Rollout consistency:} Probability \textbf{all 4 of 4 rollouts} on the same prompt are correct. \\
    \textbf{4} & \textbf{family pass@4} (4-prompt family) & $1 - \prod_j C(n-c_j, 4) / C(n, 4)$ & \textbf{Family capability ceiling:} Probability \textbf{at least 1 of 16 rollouts} across the family is correct. \\
    \textbf{5} & \textbf{family pass\hfecaret{}4} (4-prompt family) & $\prod_j C(c_j, 4) / C(n, 4)$ & \textbf{Primary floor metric:} Probability \textbf{all 16 rollouts} (4 per prompt) across the family are correct. \\
    \textbf{6} & \textbf{Family Reliability Gap} (Ceiling $-$ Floor) & family pass@4 $-$ family pass\hfecaret{}4 & \textbf{Main metric:} How often a model that can solve a family fails to solve it reliably. \\
    \bottomrule
  \end{tabularx}
\end{table}

\subsection{Grading}
\label{sec:grading}

The final answer of each model response is graded against its ground-truth reference, consisting of either a golden answer or a rubric definition, by a deterministic LLM judge (Gemini~3.1 Pro, temperature~0). Internal reasoning traces and control tokens are stripped prior to grading, and empty visible outputs score~0. Given the prompt, stripped response, and ground-truth reference, the judge returns a binary \emph{Yes}/\emph{No}. The grading prompt specifies equivalence rules for choice formatting, numeric expressions, temporal units, and multi-criterion rubrics (provided in full in the Appendix). In a cross-provider audit of all 1,492 failed prompt--model pairs across Table~\ref{tab:main_results} using Claude Fable~5.2 (weighted across 5,552 wrong-answer rollouts; \S\ref{sec:failure-modes}, Figure~\ref{fig:wrong_answers}A), only 3\% of wrong-answer rollouts ($\sim$0.3\% of all 58,752 rollouts) were flagged as potentially correct or defensible, indicating a low false-negative rate.

\subsection{Experiment Setup}
\label{sec:setup}

Models are evaluated zero-shot in a single turn with no system prompt and no access to external tools (search, code execution). Multimodal prompts pass raw, uncropped image bytes. For each of the 544 prompts, we sample $n = 6$ rollouts (3,264 per model) using provider-default decoding parameters (e.g., temperature and top-$p$) at the reasoning effort levels specified in Tables~\ref{tab:main_results} and~\ref{tab:full_results}, scoring failed or empty generations as~0.

\section{Results}
\label{sec:results}

\subsection{Overall Performance}
\label{sec:overall-performance}

\begin{figure}[!htbp]
  \centering
  \includegraphics[width=0.86\textwidth]{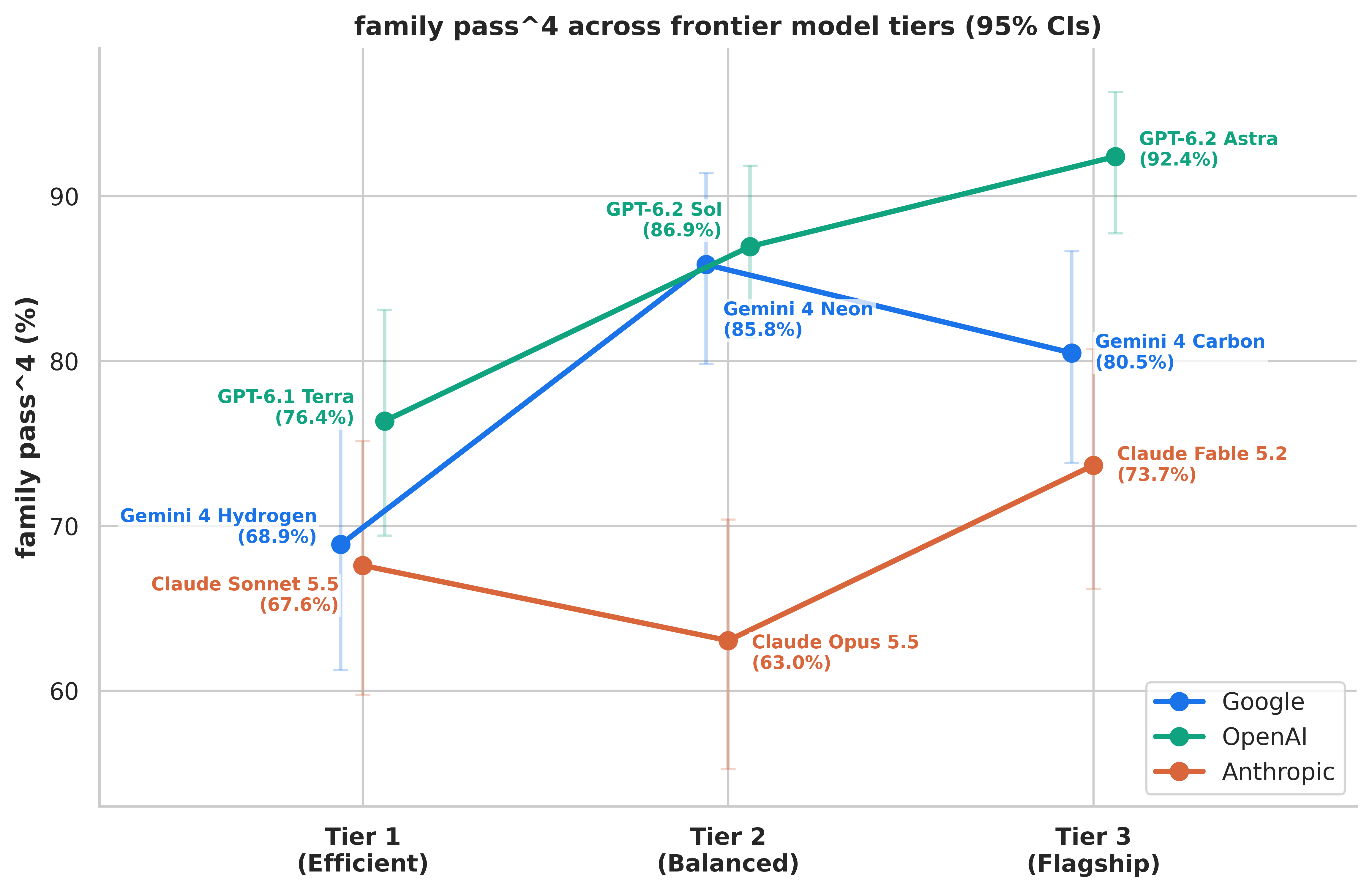}
  \caption{Family pass\hfecaret{}4 (with 95\% CIs) across the three latest frontier-family tiers per provider (Tier~1: Efficient, Tier~2: Balanced, and Tier~3: Flagship) at highest evaluated reasoning effort, excluding fast/lite models.}
  \label{fig:model_tiers}
\end{figure}

Evaluating 18 models on HFE, we observe the following:
\begin{enumerate}[leftmargin=*, itemsep=2pt]
  \item \textbf{Frontier models nearly saturate the benchmark when asked to succeed at least once.} The top five models reach 96\% to 98\% pass@1 and 97\% to 98\% pass@4, and for every model in the top~12, family pass@4 is 99.3\% to 100\% (Table~\ref{tab:main_results}).
  \item \textbf{However, they still have substantial headroom when asked to succeed every time.} Family pass\hfecaret{}4 maxes out at 92.4\% (GPT-6.2 Astra), followed by 86.9\% for GPT-6.2 Sol and 85.8\% for Gemini~4 Neon. Because family pass@4 is close to 100\% for every model, the Family Reliability Gap is almost entirely driven by family pass\hfecaret{}4, and ranges from 7.6~pp to 80.1~pp (Figure~\ref{fig:hero_gap}; complete 18-model breakdown in Appendix Figure~\ref{fig:reliability_gap_full18}).
  \item \textbf{Small differences on pass@1 translate into large gaps when consistency is required.} For instance, Claude Opus~5.5 trails GPT-6.2 Astra by 6.6~pp on pass@1, yet by 29.4~pp on family pass\hfecaret{}4. This widening is expected because any failure among the 16 sampled rollouts fails the family, illustrating how quickly small per-prompt error rates compound when consistent execution across repeated runs and minor prompt variations is required. On average across models, roughly half of the family pass\hfecaret{}4 shortfall comes from consistent prompt failures (0 of 6 rollouts correct) and half from run-to-run flakiness (between 1 and 5 of 6 correct).
  \item \textbf{Floor reliability generally scales with model tier within each provider,} though not monotonically for Google or Anthropic models (Figure~\ref{fig:model_tiers}): for Google, the balanced-tier Gemini~4 Neon (85.8\%) outperforms the flagship Gemini~4 Carbon (80.5\%), while for Anthropic, Claude Opus~5.5 trails Claude Sonnet~5.5, largely due to refusals (\S\ref{sec:failure-modes}), before recovering on Claude Fable~5.2.
\end{enumerate}

\begin{table}[!htbp]
  \centering
  \caption{HFE performance across 18 frontier and fast-tier models ($n=544$ prompts, 136 families; 11 additional reference models appear in Appendix Table~\ref{tab:full_results}).}
  \label{tab:main_results}
  \resizebox{\textwidth}{!}{%
  \renewcommand{\arraystretch}{1.15}
  \begin{tabular}{@{}r l l c c c c c c r@{}}
    \toprule
    \textbf{Rank} & \textbf{Model} & \textbf{Effort} & \thead{\textbf{pass@1}\\\textbf{($\pm$95\% CI)}} & \thead{\textbf{pass@4}\\\textbf{($\pm$95\% CI)}} & \thead{\textbf{pass\hfecaret{}4 / Rel@4}\\\textbf{($\pm$95\% CI)}} & \thead{\textbf{family pass@4}\\\textbf{(4$\times$4, 95\% Wilson CI)}} & \thead{\textbf{family pass\hfecaret{}4}\\\textbf{(4$\times$4, $\pm$95\% CI)}} & \thead{\textbf{Family Reliability Gap}\\\textbf{(fam pass@4 $-$ fam pass\hfecaret{}4)}} & \thead{\textbf{Avg Thought}\\\textbf{Tokens}} \\
    \midrule
    1 & \textbf{GPT-6.2 Astra} & xhigh & 97.8\% ($\pm$1.5) & 97.9\% ($\pm$1.5) & 97.5\% ($\pm$1.6) & 100.0\% (97.3, 100.0) & 92.4\% ($\pm$4.3) & 7.6 pp ($\pm$4.3) & 204 \\
    2 & \textbf{GPT-6.2 Sol} & xhigh & 97.3\% ($\pm$1.5) & 98.4\% ($\pm$1.2) & 95.8\% ($\pm$1.9) & 100.0\% (97.3, 100.0) & 86.9\% ($\pm$5.2) & 13.1 pp ($\pm$5.2) & 274 \\
    3 & \textbf{Gemini 4 Neon} & xhigh & 96.8\% ($\pm$1.5) & 97.9\% ($\pm$1.3) & 95.6\% ($\pm$1.9) & 100.0\% (97.3, 100.0) & 85.8\% ($\pm$5.8) & 14.2 pp ($\pm$5.8) & 9,852 \\
    4 & \textbf{GPT-6 Astra} & xhigh & 96.8\% ($\pm$1.6) & 98.0\% ($\pm$1.4) & 94.9\% ($\pm$2.0) & 100.0\% (97.3, 100.0) & 83.4\% ($\pm$5.8) & 16.6 pp ($\pm$5.8) & 382 \\
    5 & \textbf{Gemini 4 Carbon} & high & 95.6\% ($\pm$1.7) & 97.0\% ($\pm$1.3) & 93.8\% ($\pm$2.2) & 100.0\% (97.3, 100.0) & 80.5\% ($\pm$6.4) & 19.5 pp ($\pm$6.4) & 6,616 \\
    6 & \textbf{GPT-6.1 Sol} & xhigh & 96.1\% ($\pm$1.7) & 97.8\% ($\pm$1.6) & 93.7\% ($\pm$2.2) & 100.0\% (97.3, 100.0) & 79.1\% ($\pm$6.4) & 20.9 pp ($\pm$6.4) & 396 \\
    7 & \textbf{Gemini 4 Argon} & high & 94.1\% ($\pm$2.1) & 96.4\% ($\pm$1.6) & 91.5\% ($\pm$2.8) & 100.0\% (97.3, 100.0) & 76.6\% ($\pm$6.8) & 23.4 pp ($\pm$6.8) & 1,838 \\
    8 & \textbf{GPT-6.1 Terra} & xhigh & 94.7\% ($\pm$2.1) & 96.6\% ($\pm$1.9) & 91.8\% ($\pm$2.7) & 100.0\% (97.3, 100.0) & 76.4\% ($\pm$6.9) & 23.6 pp ($\pm$6.9) & 622 \\
    9 & \textbf{Claude Fable 5.2} & max & 93.4\% ($\pm$2.2) & 95.3\% ($\pm$1.9) & 91.5\% ($\pm$2.6) & 100.0\% (97.3, 100.0) & 73.7\% ($\pm$7.3) & 26.3 pp ($\pm$7.3) & 2,951 \\
    10 & \textbf{Gemini 4 Hydrogen} & xhigh & 92.5\% ($\pm$2.4) & 95.6\% ($\pm$1.7) & 88.4\% ($\pm$3.2) & 100.0\% (97.3, 100.0) & 68.9\% ($\pm$7.5) & 31.1 pp ($\pm$7.5) & 13,775 \\
    11 & \textbf{Claude Sonnet 5.5} & max & 91.7\% ($\pm$2.4) & 94.5\% ($\pm$1.9) & 89.2\% ($\pm$2.9) & 100.0\% (97.3, 100.0) & 67.6\% ($\pm$7.7) & 32.4 pp ($\pm$7.7) & 6,340 \\
    12 & \textbf{Claude Opus 5.5} & max & 91.2\% ($\pm$2.6) & 94.2\% ($\pm$2.3) & 87.8\% ($\pm$3.0) & 99.3\% (96.0, 99.9) & 63.0\% ($\pm$7.6) & 36.2 pp ($\pm$7.5) & 5,162 \\
    13 & \textbf{Gemini 3.8 Flash} & high & 87.1\% ($\pm$3.2) & 91.9\% ($\pm$2.6) & 81.7\% ($\pm$4.0) & 100.0\% (97.3, 100.0) & 56.8\% ($\pm$8.1) & 43.2 pp ($\pm$8.1) & 8,150 \\
    14 & \textbf{Claude Fable 5.1} & max & 85.7\% ($\pm$3.1) & 89.9\% ($\pm$2.7) & 80.7\% ($\pm$3.8) & 100.0\% (97.3, 100.0) & 49.7\% ($\pm$8.0) & 50.3 pp ($\pm$8.0) & 4,665 \\
    15 & \textbf{Gemini 3.6 Flash} & high & 83.8\% ($\pm$3.4) & 90.9\% ($\pm$2.7) & 75.8\% ($\pm$4.4) & 100.0\% (97.3, 100.0) & 44.7\% ($\pm$8.0) & 55.3 pp ($\pm$8.0) & 3,473 \\
    16 & \textbf{Claude Haiku 5.5} & max & 83.6\% ($\pm$3.3) & 89.4\% ($\pm$2.8) & 76.3\% ($\pm$4.0) & 99.8\% (96.8, 100.0) & 41.6\% ($\pm$7.9) & 58.1 pp ($\pm$7.9) & 13,351 \\
    17 & \textbf{GPT-6 Luna} & xhigh & 74.8\% ($\pm$4.1) & 85.4\% ($\pm$3.7) & 60.9\% ($\pm$5.0) & 98.5\% (94.8, 99.6) & 25.6\% ($\pm$6.7) & 73.0 pp ($\pm$6.9) & 1,545 \\
    18 & \textbf{Gemini 3.5 Flash-Lite} & high & 68.7\% ($\pm$4.4) & 83.6\% ($\pm$3.6) & 52.6\% ($\pm$5.3) & 98.8\% (95.1, 99.7) & 18.6\% ($\pm$6.0) & 80.1 pp ($\pm$6.1) & 3,094 \\
    \bottomrule
  \end{tabular}%
  }
  \tablenotes{Family-level metrics (family pass@4, Family Reliability Gap, and family pass\hfecaret{}4) are averaged across all four-prompt families with 95\% confidence intervals computed via family-clustered bootstrap, except family pass@4, which uses a Wilson score interval (to avoid degenerate confidence intervals where family pass@4 is 100.0\% in every bootstrap sample). Family pass@4 reaches 100\% for most models because in essentially every family, at least one prompt is solved on 3 or more of its 6 rollouts, so the empirical probability that all 16 draws fail is zero. Differences are calculated from unrounded values. Thought-token counts reflect each provider's native tokenizer and API reporting (e.g., OpenAI \texttt{reasoning\_tokens}; Gemini~4 Argon low in Tables~\ref{tab:full_results} and~\ref{tab:effort} reports 0 thought tokens) and are not directly comparable across providers. Models are ordered by descending family pass\hfecaret{}4 at highest evaluated reasoning effort.}
\end{table}

\subsection{Performance Disaggregated by Task Categories and Modality}
\label{sec:disaggregated}

\begin{figure}[!htbp]
  \centering
  \includegraphics[width=0.94\textwidth, trim=0 2pt 0 0, clip]{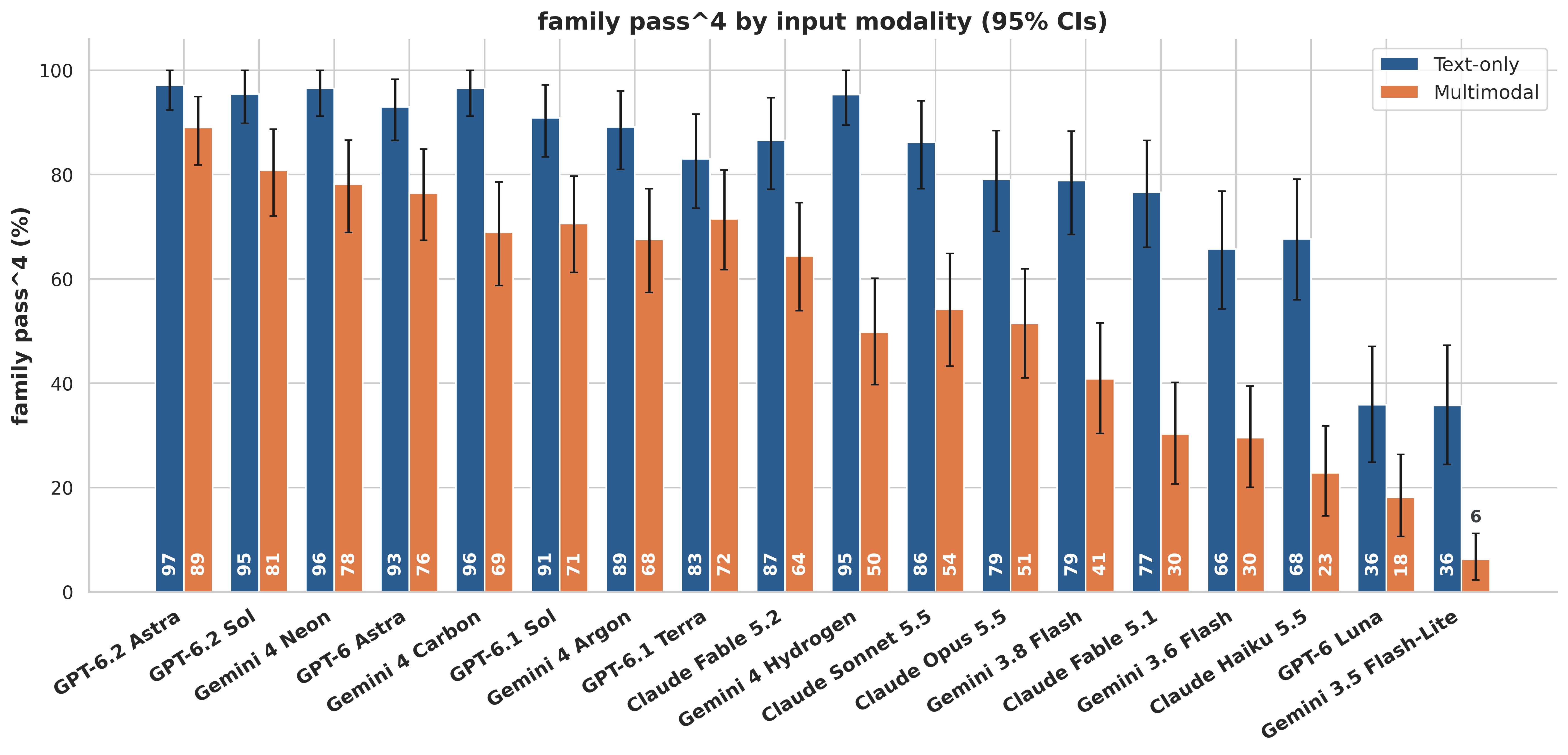}
  \caption{Family pass\hfecaret{}4 disaggregated by input modality.}
  \label{fig:modality}
\end{figure}

To isolate the drivers of floor reliability, we disaggregate family pass\hfecaret{}4 performance across four macro-categories: \textbf{Math \& Algorithmic} ($n=40$ families), \textbf{Reasoning \& Language} ($n=18$ families), \textbf{Spatial \& Positional} ($n=27$ families), and \textbf{Vision} ($n=51$ families) (Table~\ref{tab:macro_categories}) as well as by input modality (\textbf{Text-only} vs.\ \textbf{Multimodal}; Figure~\ref{fig:modality}). We find the following:
\begin{enumerate}[leftmargin=*, itemsep=2pt]
  \item \textbf{Multimodal tasks are a major source of reliability failures across all model classes.} While the top~9 models reach 83\% to 97\% family pass\hfecaret{}4 on text-only families, their performance drops to 64\% to 89\% on multimodal families (Figure~\ref{fig:modality}).
  \item \textbf{The size of this multimodal drop differs sharply across models.} It is smallest for GPT-6.2 Astra (8.0~pp) and GPT-6.1 Terra (11.5~pp), and largest for Claude Haiku~5.5, Gemini~4 Hydrogen, and Claude Fable~5.1, whose family pass\hfecaret{}4 drops by 44.8, 45.5, and 46.4 percentage points, respectively.
  \item \textbf{Frontier separation concentrates in Vision and Spatial \& Positional tasks.} Across the top~9 models in Table~\ref{tab:macro_categories}, scores span 15.0~pp on Math \& Algorithmic (83.3\% to 98.3\%, led by GPT-6.2 Astra at 98.3\% and Gemini~4 Neon and Gemini~4 Carbon at 97.5\%) and 14.7~pp on Reasoning \& Language (74.2\% to 88.9\%, with GPT-6.2 Astra, GPT-6 Astra, and Gemini~4 Argon tied at 88.9\%), compared with 18.5~pp on Spatial \& Positional (71.6\% to 90.1\%) and 27.8~pp on Vision (62.4\% to 90.2\%).
\end{enumerate}

\begin{table}[!htbp]
  \centering
  \caption{Family pass\hfecaret{}4 disaggregated by macro-category (95\% CIs in parentheses; best in each column in bold).}
  \label{tab:macro_categories}
  \resizebox{\textwidth}{!}{%
  \renewcommand{\arraystretch}{1.15}
  \begin{tabular}{@{}r l l c c c c@{}}
    \toprule
    \textbf{Rank} & \textbf{Model} & \textbf{Effort} & \thead{\textbf{Math \& Algorithmic}\\\textbf{($N=160$)}} & \thead{\textbf{Reasoning \& Language}\\\textbf{($N=72$)}} & \thead{\textbf{Spatial \& Positional}\\\textbf{($N=108$)}} & \thead{\textbf{Vision}\\\textbf{($N=204$)}} \\
    \midrule
    1 & \textbf{GPT-6.2 Astra} & xhigh & \textbf{98.3\%} (95.0, 100.0) & \textbf{88.9\%} (72.2, 100.0) & \textbf{90.1\%} (77.8, 100.0) & \textbf{90.2\%} (82.4, 98.0) \\
    2 & \textbf{GPT-6.2 Sol} & xhigh & 96.0\% (89.7, 100.0) & 83.7\% (67.0, 100.0) & 85.6\% (71.2, 96.7) & 81.7\% (71.2, 90.8) \\
    3 & \textbf{Gemini 4 Neon} & xhigh & 97.5\% (92.5, 100.0) & 83.3\% (66.7, 100.0) & 86.4\% (72.8, 97.5) & 77.3\% (65.4, 88.2) \\
    4 & \textbf{GPT-6 Astra} & xhigh & 92.5\% (85.0, 98.3) & \textbf{88.9\%} (72.2, 100.0) & 84.0\% (71.6, 95.1) & 73.9\% (62.0, 85.0) \\
    5 & \textbf{Gemini 4 Carbon} & high & 97.5\% (92.5, 100.0) & 83.7\% (67.0, 100.0) & 73.1\% (56.8, 87.9) & 69.9\% (56.9, 81.7) \\
    6 & \textbf{GPT-6.1 Sol} & xhigh & 93.3\% (85.0, 100.0) & 80.4\% (62.2, 94.8) & 78.4\% (63.4, 91.4) & 67.9\% (55.6, 79.9) \\
    7 & \textbf{Gemini 4 Argon} & high & 87.0\% (76.7, 95.8) & \textbf{88.9\%} (72.2, 100.0) & 71.6\% (54.3, 86.4) & 66.7\% (54.3, 79.1) \\
    8 & \textbf{GPT-6.1 Terra} & xhigh & 86.6\% (76.1, 95.8) & 74.2\% (52.2, 92.6) & 72.0\% (55.6, 86.8) & 71.4\% (59.5, 82.5) \\
    9 & \textbf{Claude Fable 5.2} & max & 83.3\% (70.8, 93.3) & 83.3\% (66.7, 100.0) & 74.3\% (56.0, 89.1) & 62.4\% (49.0, 75.3) \\
    10 & \textbf{Gemini 4 Hydrogen} & xhigh & 93.3\% (85.0, 100.0) & 83.7\% (66.7, 100.0) & 62.4\% (44.9, 79.3) & 47.9\% (34.9, 60.6) \\
    11 & \textbf{Claude Sonnet 5.5} & max & 85.3\% (74.5, 94.3) & 78.1\% (56.7, 94.4) & 56.8\% (38.3, 75.3) & 55.7\% (42.0, 68.8) \\
    12 & \textbf{Claude Opus 5.5} & max & 75.2\% (62.7, 86.7) & 79.6\% (61.1, 94.4) & 55.8\% (38.5, 73.1) & 51.5\% (38.7, 64.7) \\
    13 & \textbf{Gemini 3.8 Flash} & high & 76.8\% (64.3, 88.3) & 69.6\% (48.1, 89.3) & 46.4\% (28.6, 64.7) & 42.0\% (28.8, 55.6) \\
    14 & \textbf{Claude Fable 5.1} & max & 70.8\% (57.5, 83.3) & 81.5\% (64.8, 96.3) & 33.4\% (15.1, 51.9) & 30.5\% (18.7, 42.9) \\
    15 & \textbf{Gemini 3.6 Flash} & high & 69.6\% (56.0, 82.3) & 48.5\% (27.3, 69.8) & 34.6\% (17.3, 51.9) & 29.2\% (17.6, 41.5) \\
    16 & \textbf{Claude Haiku 5.5} & max & 70.3\% (56.4, 83.3) & 37.8\% (17.8, 58.5) & 31.0\% (14.9, 48.3) & 26.1\% (15.7, 37.2) \\
    17 & \textbf{GPT-6 Luna} & xhigh & 30.0\% (18.6, 43.3) & 52.2\% (31.9, 72.6) & 9.2\% (0.7, 20.2) & 21.3\% (10.9, 32.7) \\
    18 & \textbf{Gemini 3.5 Flash-Lite} & high & 46.5\% (32.5, 60.6) & 17.2\% (0.7, 34.2) & 5.7\% (0.4, 14.1) & 4.1\% (0.8, 8.8) \\
    \bottomrule
  \end{tabular}%
  }
\end{table}

\subsection{Impact of Thinking Levels (Reasoning Effort) on Floor Reliability}
\label{sec:thinking-levels}

\begin{figure}[!htbp]
  \centering
  \includegraphics[width=\textwidth]{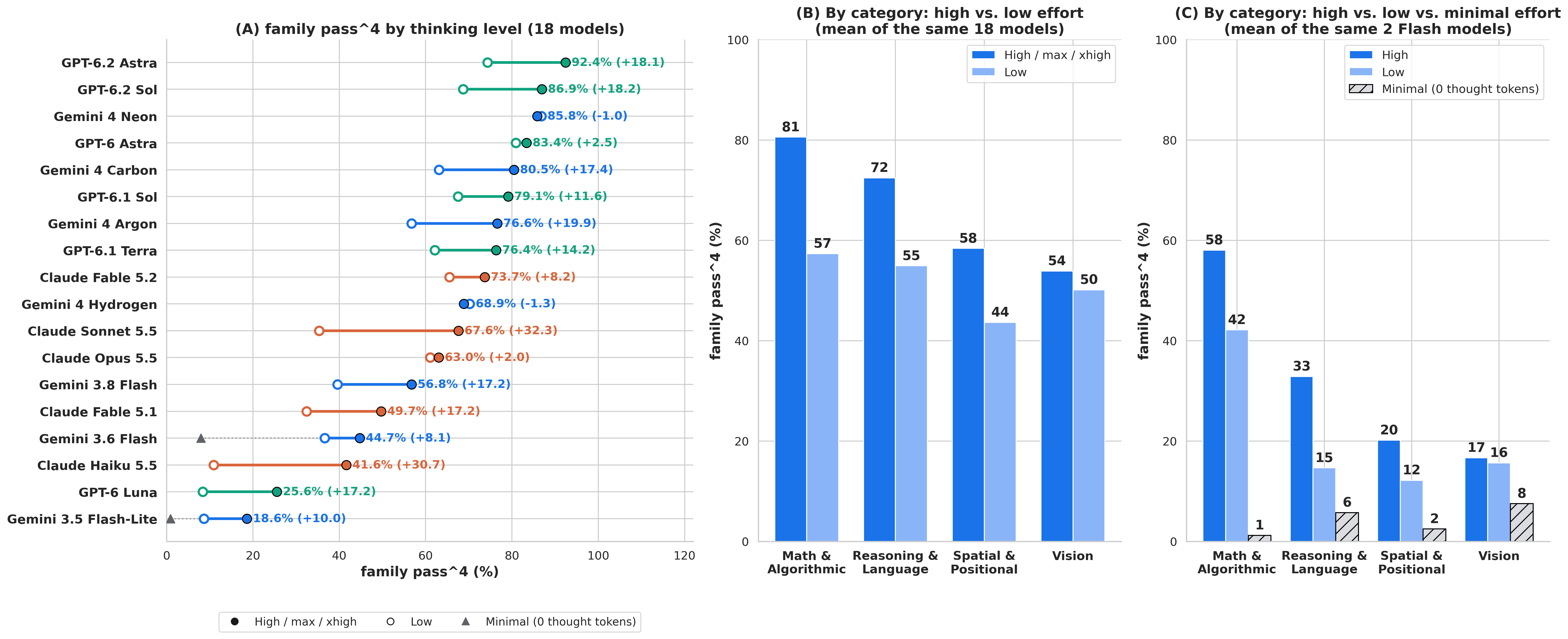}
  \caption{Thinking levels across the 18 models from Table~\ref{tab:main_results}. (A)~Family pass\hfecaret{}4 per model at each effort level. (B)~High vs.\ low effort by category, averaged over the same 18 models. (C)~High vs.\ low vs.\ minimal effort by category, averaged over the two Flash models that were run at minimal effort.}
  \label{fig:thinking}
\end{figure}

Because HFE tasks are intentionally simple for humans, a natural question is whether test-time reasoning (``thinking'') is necessary to achieve floor reliability, or whether models can solve these tasks in a single fast forward pass. To answer this, we evaluate 21 models across multiple reasoning effort configurations (Table~\ref{tab:effort}) and find:
\begin{enumerate}[leftmargin=*, itemsep=2pt]
  \item \textbf{Increasing thinking effort primarily raises the reliability floor} (family pass\hfecaret{}4) rather than the capability ceiling (family pass@4). Family pass@4 stays close to 100\% at both low and high effort, while family pass\hfecaret{}4 rises by 17.4~pp on Gemini~4 Carbon (63.1\% to 80.5\%) and 32.3~pp on Claude Sonnet~5.5 (35.3\% to 67.6\%; Figure~\ref{fig:thinking}A).
  \item \textbf{Disabling thinking causes a sharp drop in performance, but even a small thinking budget recovers most of the gain.} On Gemini~3.6 Flash, minimal effort (0 thought tokens) yields only 8.0\% family pass\hfecaret{}4, compared to 44.7\% at high effort, with low effort already capturing 78\% of that improvement (Figure~\ref{fig:thinking}A; macro-category averages across both Flash models in Figure~\ref{fig:thinking}C).
  \item \textbf{Thinking benefits symbolic and text tasks far more than visual perception.} On Gemini~4 Carbon, reducing effort from high to low drops family pass\hfecaret{}4 by 42.4~pp on Math \& Algorithmic tasks (97.5\% to 55.1\%), compared to just 3.1~pp on Vision tasks (69.9\% to 66.8\%; Table~\ref{tab:effort}; 18-model category averages in Figure~\ref{fig:thinking}B; subcategory breakdown in Table~\ref{tab:subcat_effort}).
  \item \textbf{Sensitivity to thinking effort varies widely, even between similar models.} At low effort, Gemini~4 Neon and Gemini~4 Hydrogen are statistically flat ($+1.0$ and $+1.3$~pp vs.\ xhigh, within sampling noise) and GPT-6 Astra loses only 2.5~pp family pass\hfecaret{}4, whereas GPT-6.2 Astra, GPT-6.2 Sol, and Gemini~4 Argon lose 18.1 to 19.9~pp.
\end{enumerate}

\subsection{Seed vs.\ Variant Consistency}
\label{sec:seed-vs-variant}

\begin{figure}[!htb]
  \centering
  \includegraphics[width=0.84\textwidth]{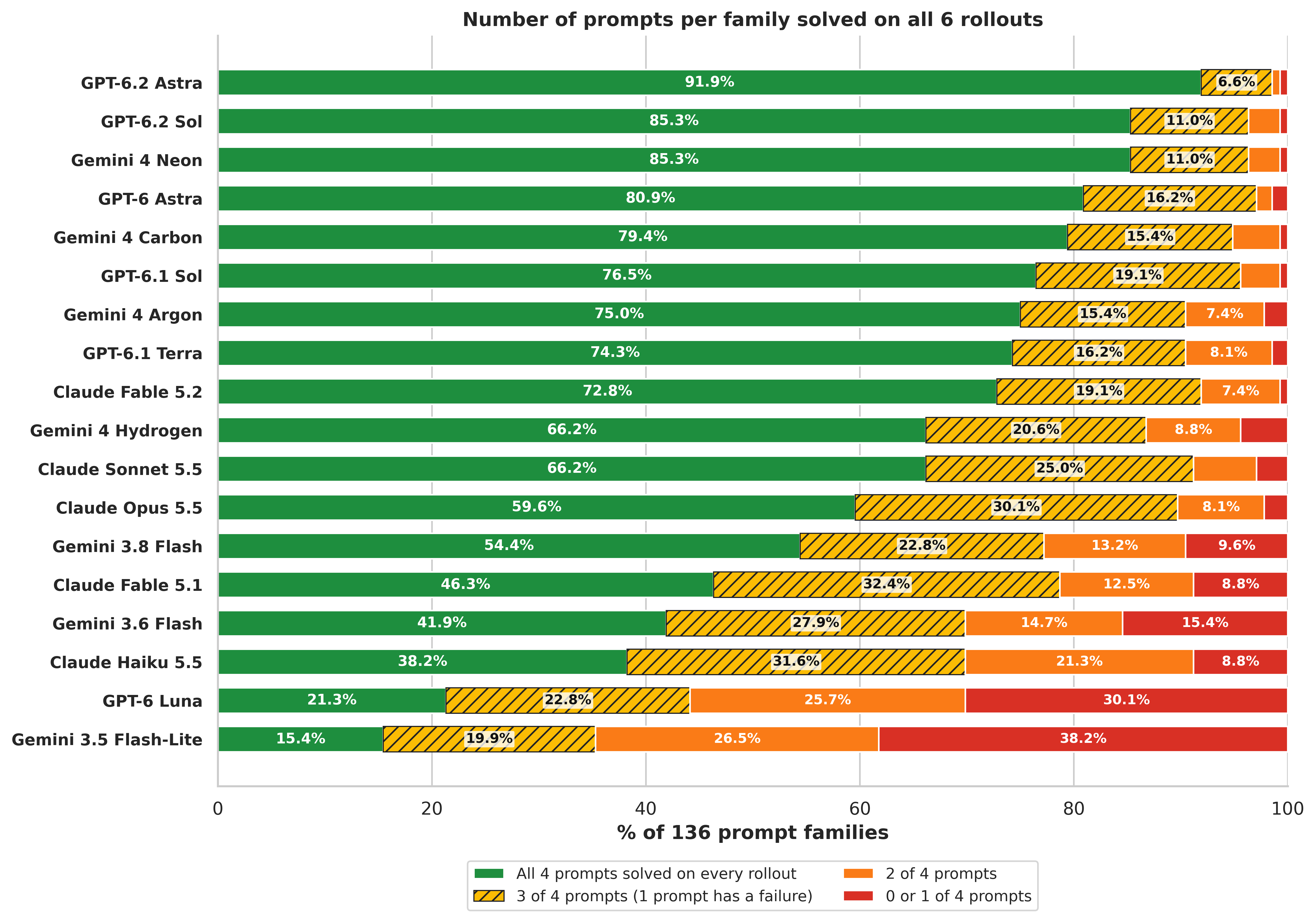}
  \caption{Share of the 136 task families with all 4, 3, 2, or $\le 1$ prompts solved across all 6 rollouts per model.}
  \label{fig:seed_vs_variant}
\end{figure}

To test whether floor failures reflect a missing capability or prompt phrasing, we look at how failures are distributed across the four prompts in each family, first within a single model and then across models. We observe the following:
\begin{enumerate}[leftmargin=*, itemsep=2pt]
  \item \textbf{Seeds and variants are similarly difficult.} Across the top~12 models, seed and variant pass\hfecaret{}4 differ at most by 4.5~pp (e.g., 95.6\% vs.\ 93.2\% on Gemini~4 Carbon), and the seed prompt is the most-failed prompt in about a quarter of failing families, matching the 25\% expected by chance.
  \item \textbf{Within a model, failing one prompt only weakly predicts failure on similar prompts.} If failures stemmed from a missing capability, a model that fails one prompt would tend to fail other prompts in the same family, and family pass\hfecaret{}4 would sit well above the independent-failure baseline ((pass\hfecaret{}4)\textsuperscript{4}). Instead, it is only 1.9 to 7.8~pp higher across the top~12 models (e.g., 80.5\% vs.\ 77.4\% on Gemini~4 Carbon; within-family $\mathrm{ICC} = 0.09\text{ to }0.20$).
  \item \textbf{Most failing families miss by a single prompt.} Among the top~12 models, 61\% to 85\% of families with any failed rollout have exactly one prompt with a failure, while the other three prompts are solved on all 6 rollouts (Figure~\ref{fig:seed_vs_variant}).
  \item \textbf{Across models, failures concentrate on the same prompt.} In 83 of the 122 families where at least two of the 18 models fail (Table~\ref{tab:main_results}), a single prompt accounts for at least half of all failed rollouts. In families where at least three models fail, the most-failed prompt receives 61\% of the failed rollouts on average, compared to 43\% if each model's failures fell on a random prompt of the family. For example, in one spatial reasoning family that passed manual QA for consistent task logic and unambiguous ground truth (\S\ref{sec:subsampling}), 15 of the 18 models fail the same variant (78 of 108 rollouts), while the other three prompts account for only 11 failed rollouts in total.
\end{enumerate}

\subsection{Human vs.\ Model Comparison}
\label{sec:human-vs-model}

\begin{figure}[!htbp]
  \centering
  \includegraphics[width=0.98\textwidth]{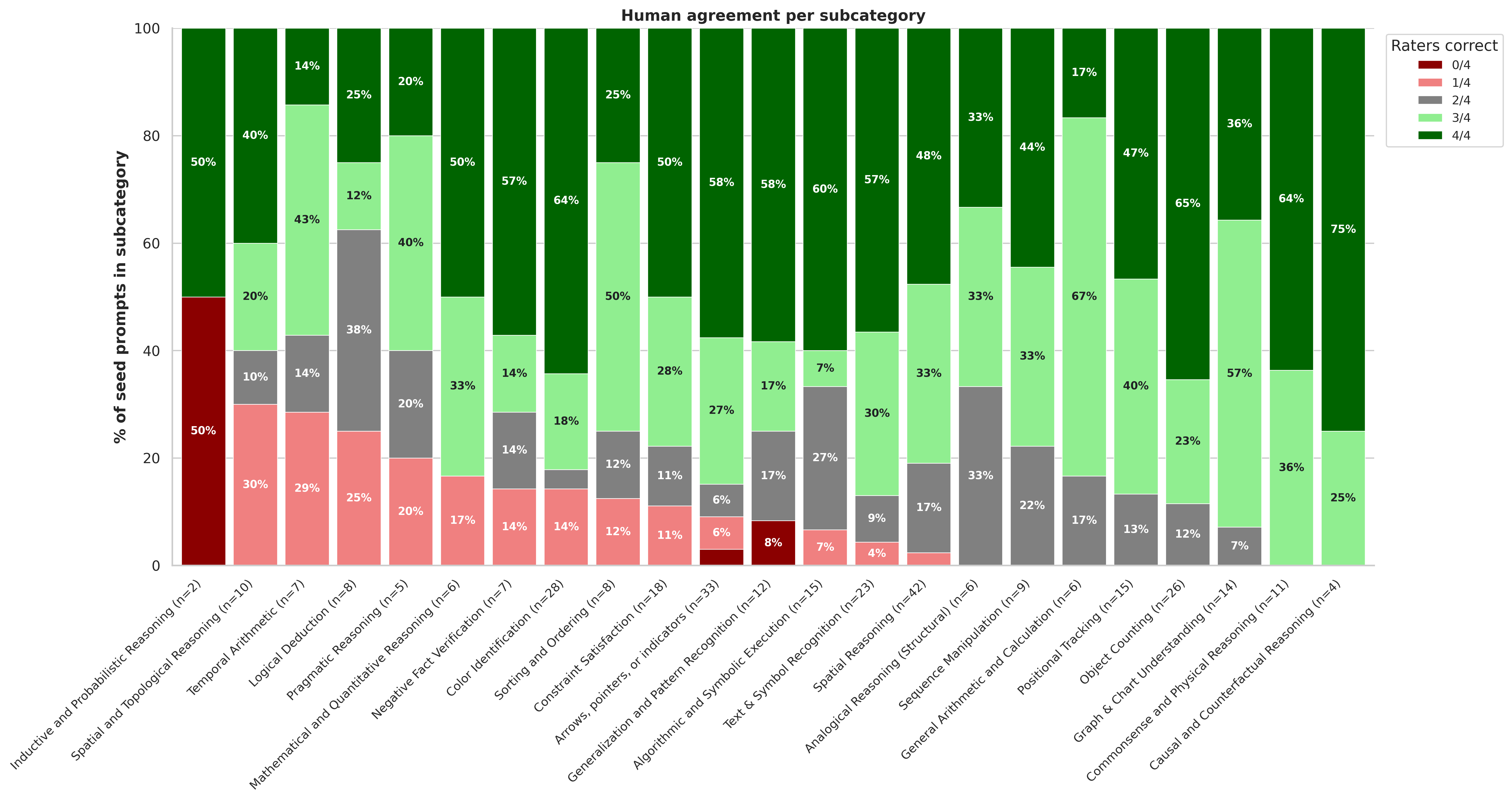}
  \caption{Human agreement per subcategory on the 315 candidate seed prompts: share of seeds solved by 0, 1, 2, 3, or 4 of the 4 raters ($n$ = number of seeds). Seeds below 3/4 were removed (\S\ref{sec:subsampling}).}
  \label{fig:human_agreement}
\end{figure}

To examine discrepancies between human and model error patterns, we compare single-rater accuracy on all 315 candidate seed prompts (\S\ref{sec:human-baseline}, before prompts that fewer than 3 of 4 raters solved were removed) with pass@1 of the 18 models on the seed prompts of the 136 final families. Because these are different prompt sets (and all candidate seeds were written to make models fail), only the ordering of categories is comparable, not the absolute accuracy. We observe the following:
\begin{enumerate}[leftmargin=*, itemsep=2pt]
  \item \textbf{Humans and models find different categories difficult.} Humans are least accurate on Math \& Algorithmic and Reasoning \& Language seeds (76\% and 77\%) and most accurate on Vision seeds (84\%). Models show the reverse order, from 98\% on Math \& Algorithmic to 83\% on Vision, and the five models used during authoring show the same order across all 315 candidate seeds.
  \item \textbf{The same reversal holds at the subcategory level.} Across the 19 subcategories with at least five candidate seeds, human and model accuracy are negatively correlated (Spearman $\rho = -0.64$; Figure~\ref{fig:human_agreement} shows human agreement by subcategory). Humans are least accurate on logical deduction, which models almost always solve, whereas models are among the least accurate on object counting and reading text in images, which are among the easiest subcategories for humans.
  \item \textbf{Human errors reflect rater mistakes rather than flaws in the prompt.} The reviewer attributed 244 of the 248 incorrect human responses to human rater error rather than a prompt or ground-truth issue. For example, three of four raters named a real treaty when asked about an event that never happened, and three of four omitted or added a letter when listing the capital letters in a sentence.
  \item \textbf{Among the final seeds, human agreement does not predict model accuracy.} Seeds solved by all four raters are no easier for models than seeds solved by three (89\% vs.\ 93\% pass@1; this holds within Vision at 81\% vs.\ 88\%, Spatial \& Positional at 87\% vs.\ 93\%, and Reasoning \& Language at 92\% vs.\ 96\%, while Math \& Algorithmic is near ceiling for both at 99\% vs.\ 98\%, suggesting this is not a category-mix artifact), and on 60 of the 86 seeds that all four raters solved, at least one model fails at least one rollout. The same conclusion holds at the full-family level: restricting the entire 18-model evaluation to the 86 families whose seed all four raters solved leaves the Family Reliability Gap virtually unchanged at 7.0 to 80.6~pp (compared with 7.6 to 80.1~pp overall), and 13 of the 18 models actually score lower on family pass\hfecaret{}4 across these unanimous 4/4 families than across the 50 3/4 families (18-model mean 64.0\% vs.\ 66.8\%; for example, Gemini~4 Carbon scores 77.7\% vs.\ 85.3\% and Claude Opus~5.5 scores 59.0\% vs.\ 70.0\%).
\end{enumerate}

\subsection{Failure Modes}
\label{sec:failure-modes}

A rollout can fail in four ways: (1)~a \textbf{wrong answer}, where the model responds and the autorater grades the response as incorrect; (2)~a \textbf{refusal}, where the provider blocks the response with a safety stop reason or content filter; (3)~an \textbf{output-limit truncation}, where the response reaches the output-token limit before giving an answer; and (4)~an \textbf{evaluation-harness error}, such as a malformed response or an API quota error. Only the first is a model error; the other three are failures without an answer. Across the 18 models evaluated, 95\% of the 5,823 failed rollouts are wrong answers, 3\% are refusals, 2\% are output-limit truncations, and 0.1\% are harness errors (Table~\ref{tab:main_results}; Table~\ref{tab:failure_types}). We first describe the wrong answers and then the other three types.

\begin{figure}[!htbp]
  \centering
  \includegraphics[width=\textwidth]{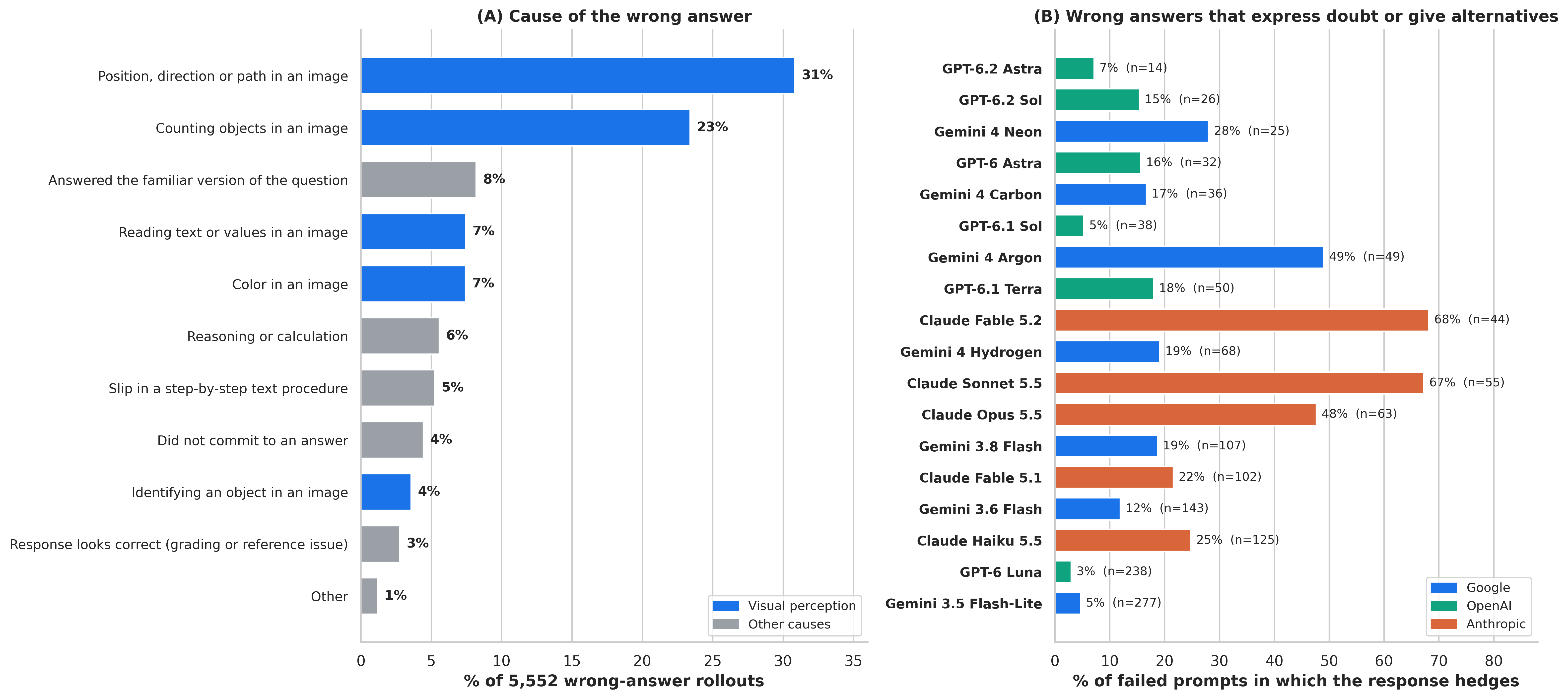}
  \caption{Behaviors behind wrong answers, as labeled by an LLM judge. (A)~Cause, as a share of wrong-answer rollouts. (B)~Share of each model's failed prompts in which the response hedges; $n$ is the number of failed prompts.}
  \label{fig:wrong_answers}
\end{figure}

\paragraph{Wrong answers (model errors).}
To characterize incorrect responses, we examined every prompt on which one of the 18 models gave at least one wrong answer (385 prompts, 1,492 prompt--model pairs, 5,552 wrong rollouts). For each pair, an LLM judge (Claude Fable~5.2, medium reasoning effort) read one failed response together with the prompt, the reference answer, and the image (if there is one), and assigned a primary cause and a set of behavior flags. Shares of wrong rollouts weight each pair by its number of wrong rollouts, which assumes the other wrong rollouts of a pair fail for the same reason. The labels were spot-checked manually but not validated by human annotators, and because the judge is itself one of the evaluated models, the shares below are approximate; a second judge cross-checked the labels. We observe the following:
\begin{enumerate}[leftmargin=*, itemsep=2pt]
  \item \textbf{About three-quarters of wrong answers are visual perception errors} (73\% of wrong rollouts, Figure~\ref{fig:wrong_answers}A). The two largest causes are misjudging a position, direction, or path in an image (31\%) and miscounting objects (23\%).
  \item \textbf{Many wrong answers are near misses.} In 40\% of failed prompt--model pairs, the answer is off by one or has a single wrong element; this holds for 173 of the 303 counting errors.
  \item \textbf{Occlusion is a recurring cause.} In 16\% of failed pairs (concentrated in 53 prompts), the error involves an object that is partly hidden, overlapping, or drawn over (for example, overlapping objects in a cluster, such as keys bunched on a ring).
  \item \textbf{Models default to answering the familiar version of a question.} In 8\% of wrong rollouts, the model solves the well-known puzzle or accepts a false premise instead of reading the changed detail (for example, giving the textbook answer to a well-known riddle whose key constraint has been altered).
  \item \textbf{Smaller models slip while executing simple text procedures.} 111 of the 132 such failures come from the smallest fast-tier models across all three providers (54 across the three Gemini Flash/Flash-Lite models, 48 from GPT-6 Luna, and 9 from Claude Haiku~5.5), and in 42 of the 132, the response derives the correct result step by step before writing a different final answer.
  \item \textbf{Hedging differs by provider} (Figure~\ref{fig:wrong_answers}B). Anthropic models express doubt or offer alternatives in 39\% of their failed pairs, compared with 14\% for Google and 7\% for OpenAI models, while declining to give any definite answer is rare (4\% of wrong rollouts). Because we did not label correct responses, we cannot say whether hedging is more common when a model is wrong.
\end{enumerate}

\paragraph{Agreement with a second judge.}
Repeating the labeling on all 1,492 failed prompt--model pairs with Gemini~4 Carbon (high reasoning effort) confirms that the primary failure taxonomy is stable: the two judges assign the same cause on 84\% of pairs (Cohen's $\kappa = 0.81$) and report nearly identical shares of visual perception errors (72\% vs.\ 73\%). Behavior-flag rates are similar or somewhat lower under the second judge (near misses 32\% vs.\ 40\%; hedging 13\% vs.\ 19\%; occlusion 16\% vs.\ 16\%), while the cross-provider hedging gap persists (26\% for Anthropic vs.\ 10\% for Google and 6\% for OpenAI).

\begin{figure}[!htbp]
  \centering
  \includegraphics[width=\textwidth]{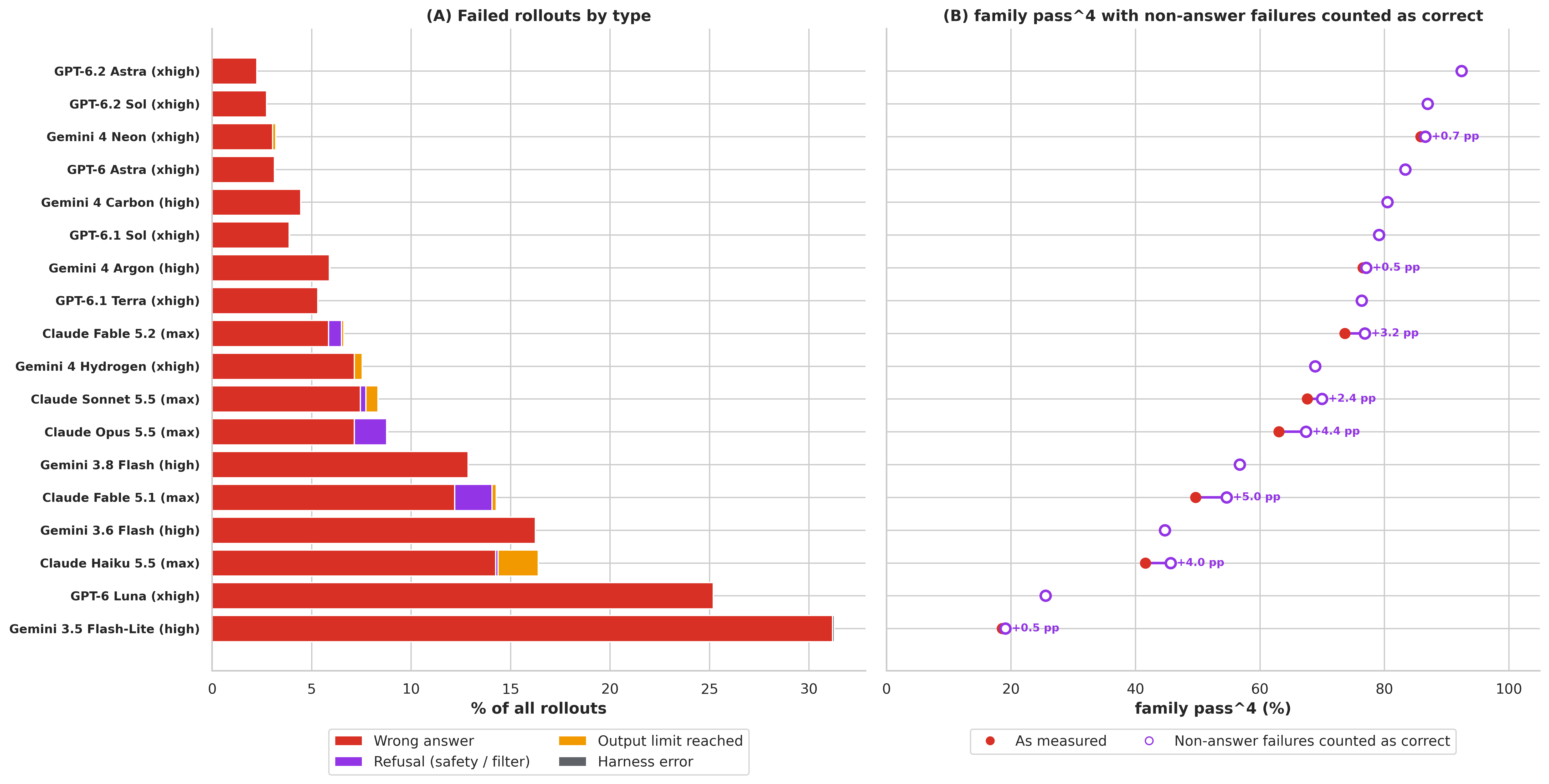}
  \caption{Failure types and their effect on family pass\hfecaret{}4.}
  \label{fig:failure_types}
\end{figure}

\paragraph{Refusals, output limits, and harness errors.}
Because family pass\hfecaret{}4 requires 16 consecutive successes, a single rollout that fails without an answer lowers the score for the whole family. We report how often each non-answer failure type occurs and how much family pass\hfecaret{}4 would change if they were counted as correct (Table~\ref{tab:failure_types}; Figure~\ref{fig:failure_types}). We observe the following:
\begin{itemize}[leftmargin=*, itemsep=2pt]
  \item \textbf{Harness errors are rare.} They occur in only 5 of 58,752 rollouts across the 18 models in Table~\ref{tab:failure_types} (and 10 of 169,728 across all 52 model runs in Table~\ref{tab:full_results}).
  \item \textbf{Refusals occur only for Anthropic models and concentrate in a few families.} Claude Opus~5.5 refuses on 1.6\% of rollouts, affecting 4.4\% of families and lowering its family pass\hfecaret{}4 by 4.4~pp (67.4\% to 63.0\%). Claude Fable~5.1 is affected similarly (1.9\% of rollouts, 4.5~pp from refusals alone), while Claude Fable~5.2 refuses less often (0.6\% of rollouts across 2.9\% of families, 2.5~pp from refusals alone). We retain refusals as failures in our primary score as we feel that models should not refuse on the benign, basic tasks in HFE.
  \item \textbf{Output-limit truncations mainly affect smaller Anthropic models with very long reasoning.} Claude Haiku~5.5 reaches the output limit on 2.0\% of rollouts, in 14.7\% of families.
  \item \textbf{Counting every refusal, output-limit truncation and harness error as correct yields the largest gains for Anthropic models, yet swaps only three adjacent pairs:} GPT-6.1 Terra (\#8 $\to$ \#9) with Claude Fable~5.2 (\#9 $\to$ \#8, $+3.2$~pp to 76.9\%), Gemini~4 Hydrogen (\#10 $\to$ \#11) with Claude Sonnet~5.5 (\#11 $\to$ \#10, $+2.4$~pp to 70.0\%), and Gemini~3.6 Flash (\#15 $\to$ \#16) with Claude Haiku~5.5 (\#16 $\to$ \#15, $+4.0$~pp to 45.6\%).
\end{itemize}

\begin{table}[!htbp]
  \centering
  \caption{Failure types per model and their effect on family pass\hfecaret{}4.}
  \label{tab:failure_types}
  \resizebox{\textwidth}{!}{%
  \renewcommand{\arraystretch}{1.15}
  \begin{tabular}{@{}l l c c c c c c c c@{}}
    \toprule
    \textbf{Model} & \textbf{Effort} & \thead{\textbf{Wrong answers}\\\textbf{(\% of rollouts)}} & \thead{\textbf{Refusals}\\\textbf{(\%)}} & \thead{\textbf{Output limit}\\\textbf{reached (\%)}} & \thead{\textbf{Harness}\\\textbf{errors (\%)}} & \thead{\textbf{Families with}\\\textbf{$\ge$1 refusal}} & \thead{\textbf{Families with $\ge$1 output-}\\\textbf{limit or harness failure}} & \textbf{family pass\hfecaret{}4} & \thead{\textbf{family pass\hfecaret{}4 if refusals, output limits}\\\textbf{and harness errors counted correct}} \\
    \midrule
    \textbf{GPT-6.2 Astra} & xhigh & 2.24\% & 0.00\% & 0.00\% & 0.00\% & 0.0\% & 0.0\% / 0.0\% & 92.4\% & 92.4\% (+0.0) \\
    \textbf{GPT-6.2 Sol} & xhigh & 2.73\% & 0.00\% & 0.00\% & 0.00\% & 0.0\% & 0.0\% / 0.0\% & 86.9\% & 86.9\% (+0.0) \\
    \textbf{Gemini 4 Neon} & xhigh & 3.03\% & 0.00\% & 0.15\% & 0.00\% & 0.0\% & 0.7\% / 0.0\% & 85.8\% & 86.6\% (+0.7) \\
    \textbf{GPT-6 Astra} & xhigh & 3.12\% & 0.00\% & 0.00\% & 0.03\% & 0.0\% & 0.0\% / 0.7\% & 83.4\% & 83.4\% (+0.0) \\
    \textbf{Gemini 4 Carbon} & high & 4.44\% & 0.00\% & 0.00\% & 0.00\% & 0.0\% & 0.0\% / 0.0\% & 80.5\% & 80.5\% (+0.0) \\
    \textbf{GPT-6.1 Sol} & xhigh & 3.86\% & 0.00\% & 0.00\% & 0.03\% & 0.0\% & 0.0\% / 0.7\% & 79.1\% & 79.2\% (+0.0) \\
    \textbf{Gemini 4 Argon} & high & 5.88\% & 0.00\% & 0.03\% & 0.00\% & 0.0\% & 0.7\% / 0.0\% & 76.6\% & 77.1\% (+0.5) \\
    \textbf{GPT-6.1 Terra} & xhigh & 5.30\% & 0.00\% & 0.00\% & 0.00\% & 0.0\% & 0.0\% / 0.0\% & 76.4\% & 76.4\% (+0.0) \\
    \textbf{Claude Fable 5.2} & max & 5.85\% & 0.64\% & 0.12\% & 0.00\% & 2.9\% & 1.5\% / 0.0\% & 73.7\% & 76.9\% (+3.2) \\
    \textbf{Gemini 4 Hydrogen} & xhigh & 7.14\% & 0.00\% & 0.40\% & 0.00\% & 0.0\% & 1.5\% / 0.0\% & 68.9\% & 68.9\% (+0.0) \\
    \textbf{Claude Sonnet 5.5} & max & 7.44\% & 0.28\% & 0.61\% & 0.00\% & 0.7\% & 4.4\% / 0.0\% & 67.6\% & 70.0\% (+2.4) \\
    \textbf{Claude Opus 5.5} & max & 7.14\% & 1.62\% & 0.06\% & 0.00\% & 4.4\% & 1.5\% / 0.0\% & 63.0\% & 67.4\% (+4.4) \\
    \textbf{Gemini 3.8 Flash} & high & 12.87\% & 0.00\% & 0.00\% & 0.00\% & 0.0\% & 0.0\% / 0.0\% & 56.8\% & 56.8\% (+0.0) \\
    \textbf{Claude Fable 5.1} & max & 12.19\% & 1.87\% & 0.21\% & 0.00\% & 5.1\% & 2.2\% / 0.0\% & 49.7\% & 54.7\% (+5.0) \\
    \textbf{Gemini 3.6 Flash} & high & 16.24\% & 0.00\% & 0.00\% & 0.00\% & 0.0\% & 0.0\% / 0.0\% & 44.7\% & 44.7\% (+0.0) \\
    \textbf{Claude Haiku 5.5} & max & 14.25\% & 0.12\% & 2.02\% & 0.00\% & 1.5\% & 14.7\% / 0.0\% & 41.6\% & 45.6\% (+4.0) \\
    \textbf{GPT-6 Luna} & xhigh & 25.18\% & 0.00\% & 0.00\% & 0.00\% & 0.0\% & 0.0\% / 0.0\% & 25.6\% & 25.6\% (+0.0) \\
    \textbf{Gemini 3.5 Flash-Lite} & high & 31.19\% & 0.00\% & 0.00\% & 0.09\% & 0.0\% & 0.0\% / 2.2\% & 18.6\% & 19.1\% (+0.5) \\
    \bottomrule
  \end{tabular}%
  }
\end{table}

\subsection{Release Subset Selection}
\label{sec:release-subset}

To prevent benchmark overfitting, we release a $\sim$20\% public subset of the dataset constructed via stratified sampling across capability categories and difficulty levels. First, we apply a Two-Parameter Logistic (2PL) Item Response Theory (IRT) model~\citep{maiapolo2024tinybenchmarks} to 84,864 evaluation outcomes, encompassing 544 prompts evaluated across an earlier release-selection pool of 26 model--effort configurations with 6 rollouts to estimate item difficulty and discriminability for each prompt.

We selected the public partition in two steps using the 136 seed prompts (excluding the three variants per family from the 544-prompt pool):
\begin{itemize}[leftmargin=*, itemsep=2pt]
  \item \textbf{Stratified Seed Sampling:} Difficulty estimates define the sampling strata per capability category. Within each stratum, we select the top 20\% of prompts ranked by discriminability score, yielding 28 representative seed prompts.
  \item \textbf{Public Set Expansion:} We include the 3 variants derived from each chosen seed prompt, completing a 112-prompt (28-family) public evaluation set while holding out the remaining 432 prompts (108 families) privately.
\end{itemize}
Consequently, the public sample subset exhibits strong alignment with the complete dataset, achieving a Spearman's rank correlation of 0.97.

\section{Related Work}
\label{sec:related}

\subsection{Capability Ceilings vs.\ Elementary Failures}
\label{sec:related-ceilings}

Frontier evaluations largely target specialized reasoning ceilings and contamination-resistant tasks~\citep{rein2023gpqa,zhou2023ifeval,jain2024livecodebench,wang2024mmlupro,white2024livebench,phan2025hle,xu2025ugmathbench}, yet high expert scores coexist with a ``jagged'' profile on trivial problems~\citep{dellacqua2023jagged,karpathy2024jagged}. Prior benchmarks have probed specific failure modes across character-level manipulation~\citep{edman2024cute}, rule induction~\citep{hua2025inductionbench}, and simple commonsense or constraint-following tasks~\citep{nezhurina2024alice,simplebench2024,allcock2026obviousbench} as well as basic visual perception~\citep{fu2024blink,li2024naturalbench,tong2024eyes}. Building on adversarial model-in-the-loop collection~\citep{nie2020anli,kiela2021dynabench}, HFE unifies text and multimodal elementary tasks across 20 human-verified subcategories.

\subsection{Perturbation Invariance and Repeated-Sampling Reliability}
\label{sec:related-invariance}

Perturbation-based benchmarks test whether models generalize across surface variations rather than memorizing specific phrasings~\citep{mirzadeh2024gsmsymbolic}, though recent studies largely perturb difficult capability-boundary problems such as competition math~\citep{srivastava2024functional,luo2026jaggedbench}. Meanwhile, repeated-sampling studies measure multi-try coverage (pass@k; \citealp{chen2021codex,brown2024monkeys}) and test-time compute scaling optimizes single-output accuracy~\citep{snell2024scaling}, rather than consistency across repeated calls. Extending multi-trial consistency (pass\hfecaret{}k; \citealp{yao2024taubench,allcock2026obviousbench}) across four-prompt task families (family pass\hfecaret{}4), HFE jointly measures rollout reliability and variant robustness on elementary tasks that humans solve easily.

\subsection{Reliability in Agentic and Real-World Knowledge Work}
\label{sec:related-agentic}

The jaggedness in model reliability we observe on elementary prompts extends to agentic tasks and real-world knowledge work. Benchmarks and deployment reports evaluating performance in these areas show that frontier models frequently fail to complete multi-step workflows end-to-end across tool use, knowledge work, and research engineering~\citep{yao2024taubench,xu2024theagentcompany,barres2025tau2bench,patwardhan2025gdpval,openai2026b_jaggedfrontier}, and long-horizon evaluations show that reliable completion drops sharply as task length grows~\citep{kwa2025metr}. On various agentic benchmarks, run-to-run consistency and model robustness to perturbations have improved at a slower pace compared to overall accuracy~\citep{rabanser2026reliability}. Whereas these studies evaluate end-to-end performance on complex agentic workflows, HFE isolates the underlying single-turn reliability floor on elementary skills where basic perceptual or procedural slips can cause downstream tasks to fail.

\section{Discussion and Conclusion}
\label{sec:discussion-conclusion}

\subsection{Discussion}
\label{sec:discussion}

Returning to the questions posed in \S\ref{sec:intro}, frontier LLMs rarely fail elementary tasks from a lack of underlying capability: top models achieve 99.3\% to 100\% family pass@4, solving nearly every task family on at least one rollout. Yet their reliability floor remains well below ceiling (family pass\hfecaret{}4 $\le 92.4\%$, leaving a 7.6 to 80.1~pp Family Reliability Gap), with most family failures triggered by a single prompt, often the same prompt across providers (\S\ref{sec:overall-performance}, \S\ref{sec:seed-vs-variant}). Scaling test-time thinking only partially closes this gap: increased reasoning effort substantially lifts family pass\hfecaret{}4 on symbolic and text tasks by catching procedural slips, yet yields small or negligible gains on Vision (\S\ref{sec:thinking-levels}), which inverts human difficulty ordering (Spearman $\rho = -0.64$; \S\ref{sec:human-vs-model}). Visual perception errors account for 73\% of wrong answers (\S\ref{sec:failure-modes}). Raising baseline reliability therefore requires stronger visual perception and robustness to surface perturbations, not test-time compute alone.

\subsection{Limitations}
\label{sec:limitations}

Four methodological choices bound the interpretation of our results. First, candidate seeds and variants were adversarially filtered to elicit at least one failure on our calibration panel during authoring (\S\ref{sec:seeds} and \S\ref{sec:families}), and outputs from evaluated models also informed the QA audit and the choice of which variant to drop (\S\ref{sec:subsampling}). HFE thus measures reliability on elementary tasks selected to expose model blind spots rather than a uniform sample of everyday queries, though 16 of the 18 evaluated models (Table~\ref{tab:main_results}), including all frontier tiers, post-date collection and perform similarly on seeds and variants (\S\ref{sec:seed-vs-variant}). Second, most multimodal prompts pair text with Gemini-generated images alongside diagrams, screenshots, and photographs. Synthetic imagery can introduce visual generation artifacts and does not span the full distribution of natural photographs, though human raters achieved their highest accuracy on Vision seeds (84\%; \S\ref{sec:human-vs-model}) and non-Google models achieve the highest multimodal family pass\hfecaret{}4 (\S\ref{sec:disaggregated}), showing no systematic advantage for Gemini models. Third, responses are graded by a single deterministic Gemini~3.1 Pro autorater (\S\ref{sec:grading}); although tasks use short, objective reference answers or rubrics and passed multi-stage quality review (\S\ref{sec:subsampling}), occasional residual prompt ambiguities, edge-case reference answers, or autorater misclassifications cannot be fully ruled out, and relying on one model family for grading could introduce subtle same-provider bias. Fourth, our human baseline evaluated seed prompts only (25\% of the final benchmark, not the three sibling variants) using single-attempt, unassisted raters rather than repeated multi-variant trials (human family pass\hfecaret{}4). While working without tools such as calculators can occasionally cause human clerical slips even on conceptually elementary tasks, \S\ref{sec:seed-vs-variant} shows that models fail seeds and variants at similar rates, and \S\ref{sec:human-vs-model} confirms that restricting HFE strictly to the 86 families with unanimous 4/4 human agreement preserves the full reliability gap (7.0 to 80.6~pp).

\subsection{Future Work}
\label{sec:future-work}

What constitutes the ``reliability floor'' is a moving target: as frontier capabilities advance, expectations of what counts as a ``basic'' model competency will rise over time, eventually including tasks that are difficult for humans, such as challenging mathematics or coding problems, where consistent reliability will increasingly be expected as routine. Tracking this rising baseline motivates three extensions of HFE:
\begin{enumerate}[leftmargin=*, itemsep=2pt]
  \item Floor evaluations should be continually refreshed (for example, through open, community-driven task collection) as yesterday's capability ceiling becomes tomorrow's expected floor.
  \item Whereas we restricted HFE to objectively gradable text outputs (\S\ref{sec:taxonomy}), extending family pass\hfecaret{}k to multimodal generation (such as image generation and editing, diagram synthesis, and audio) would test whether elementary counting, spatial, and attribute-binding failures similarly limit generative reliability.
  \item Extending floor-reliability evaluation from standalone, single-turn prompts (\S\ref{sec:setup}) to agentic tasks (including tool and computer use) would measure whether models can complete everyday multi-step workflows consistently, where a single basic mistake can derail the entire task.
\end{enumerate}

\subsection{Conclusion}
\label{sec:conclusion}

We introduced Humanity's First Exam (HFE), a benchmark of 544 text and multimodal prompts (136 four-prompt task families) for measuring the foundational reliability of LLMs on elementary tasks that humans can easily solve. Across 18 models, near-perfect multi-sample solvability coexists with substantial inconsistency across repeated rollouts and minor surface variations, concentrated in basic visual perception where extra test-time thinking offers limited relief. Trustworthy deployment depends on models succeeding every time rather than merely once. Through our family-reliability methodology, 112-prompt public subset, and held-out private evaluation, we aim to encourage practices that raise the reliability floor alongside the capability ceiling.

\Needspace{12\baselineskip}
\section*{Acknowledgments}
\noindent We thank our colleagues across Google DeepMind and Kaggle for their support and feedback throughout this project, with special thanks to:
\begin{itemize}[leftmargin=*, itemsep=2pt]
  \item \textbf{Sponsors:} Demis Hassabis, Logan Kilpatrick, and Tulsee Doshi.
  \item \textbf{Advisors:} Kate Olszewska, Apurv Suman, and Nico Duduta.
  \item \textbf{Operational, Editorial \& Release Support:} Alison~Snyder, Parth~Modi, Sean~Nakamoto, and Anand~Rao.
  \item \textbf{Public Leaderboard:} The Kaggle team for hosting the public HFE leaderboard and supporting community task contributions.
  \item \textbf{AI Assistance:} Gemini was used for coding, dataset auditing, data analysis, and drafting and editorial assistance during the preparation of this manuscript.
\end{itemize}

\clearpage
\section{Appendix}
\label{sec:appendix}
\setcounter{table}{0}
\renewcommand{\thetable}{A\arabic{table}}
\setcounter{figure}{0}
\renewcommand{\thefigure}{A\arabic{figure}}

\begin{table}[!htbp]
  \centering
  \caption{Complete HFE results across all 29 models and 52 reasoning-effort configurations ($n=544$ prompts, 136 families).}
  \label{tab:full_results}
  \resizebox{\textwidth}{!}{%
  \renewcommand{\arraystretch}{0.95}
  \begin{tabular}{@{}r l l c c c c c c r@{}}
    \toprule
    \textbf{Rank} & \textbf{Model} & \textbf{Effort} & \thead{\textbf{pass@1}\\\textbf{($\pm$95\% CI)}} & \thead{\textbf{pass@4}\\\textbf{($\pm$95\% CI)}} & \thead{\textbf{pass\hfecaret{}4 / Rel@4}\\\textbf{($\pm$95\% CI)}} & \thead{\textbf{family pass@4}\\\textbf{(4$\times$4, 95\% Wilson CI)}} & \thead{\textbf{family pass\hfecaret{}4}\\\textbf{(4$\times$4, $\pm$95\% CI)}} & \thead{\textbf{Family Reliability Gap}\\\textbf{(fam pass@4 $-$ fam pass\hfecaret{}4)}} & \thead{\textbf{Avg Thought}\\\textbf{Tokens}} \\
    \midrule
    1 & \textbf{GPT-6.2 Astra} & xhigh & 97.8\% ($\pm$1.5) & 97.9\% ($\pm$1.5) & 97.5\% ($\pm$1.6) & 100.0\% (97.3, 100.0) & 92.4\% ($\pm$4.3) & 7.6 pp ($\pm$4.3) & 204 \\
    1 & \textbf{GPT-6.2 Astra} & low & 94.9\% ($\pm$1.9) & 97.9\% ($\pm$1.5) & 90.7\% ($\pm$2.9) & 100.0\% (97.3, 100.0) & 74.3\% ($\pm$7.0) & 25.7 pp ($\pm$7.0) & 14 \\
    2 & \textbf{GPT-6.2 Sol} & xhigh & 97.3\% ($\pm$1.5) & 98.4\% ($\pm$1.2) & 95.8\% ($\pm$1.9) & 100.0\% (97.3, 100.0) & 86.9\% ($\pm$5.2) & 13.1 pp ($\pm$5.2) & 274 \\
    2 & \textbf{GPT-6.2 Sol} & low & 93.1\% ($\pm$2.5) & 96.8\% ($\pm$1.9) & 87.3\% ($\pm$3.7) & 100.0\% (97.2, 100.0) & 68.7\% ($\pm$7.1) & 31.3 pp ($\pm$7.1) & 15 \\
    3 & \textbf{Gemini 4 Neon} & xhigh & 96.8\% ($\pm$1.5) & 97.9\% ($\pm$1.3) & 95.6\% ($\pm$1.9) & 100.0\% (97.3, 100.0) & 85.8\% ($\pm$5.8) & 14.2 pp ($\pm$5.8) & 9,852 \\
    3 & \textbf{Gemini 4 Neon} & low & 96.9\% ($\pm$1.7) & 97.6\% ($\pm$1.4) & 96.0\% ($\pm$1.9) & 100.0\% (97.3, 100.0) & 86.8\% ($\pm$5.5) & 13.2 pp ($\pm$5.5) & 787 \\
    4 & \textbf{GPT-6 Astra} & xhigh & 96.8\% ($\pm$1.6) & 98.0\% ($\pm$1.4) & 94.9\% ($\pm$2.0) & 100.0\% (97.3, 100.0) & 83.4\% ($\pm$5.8) & 16.6 pp ($\pm$5.8) & 382 \\
    4 & \textbf{GPT-6 Astra} & low & 96.1\% ($\pm$1.8) & 97.9\% ($\pm$1.5) & 93.8\% ($\pm$2.3) & 100.0\% (97.3, 100.0) & 80.9\% ($\pm$6.3) & 19.1 pp ($\pm$6.3) & 65 \\
    5 & \textbf{Gemini 4 Carbon} & high & 95.6\% ($\pm$1.7) & 97.0\% ($\pm$1.3) & 93.8\% ($\pm$2.2) & 100.0\% (97.3, 100.0) & 80.5\% ($\pm$6.4) & 19.5 pp ($\pm$6.4) & 6,616 \\
    5 & \textbf{Gemini 4 Carbon} & low & 90.9\% ($\pm$2.5) & 95.6\% ($\pm$1.8) & 84.4\% ($\pm$3.9) & 100.0\% (97.3, 100.0) & 63.1\% ($\pm$7.7) & 36.9 pp ($\pm$7.7) & 29 \\
    6 & \textbf{GPT-6.1 Sol} & xhigh & 96.1\% ($\pm$1.7) & 97.8\% ($\pm$1.6) & 93.7\% ($\pm$2.2) & 100.0\% (97.3, 100.0) & 79.1\% ($\pm$6.4) & 20.9 pp ($\pm$6.4) & 396 \\
    6 & \textbf{GPT-6.1 Sol} & low & 94.0\% ($\pm$2.0) & 96.7\% ($\pm$1.7) & 90.0\% ($\pm$2.5) & 100.0\% (97.3, 100.0) & 67.5\% ($\pm$7.3) & 32.5 pp ($\pm$7.3) & 79 \\
    7 & \textbf{Gemini 4 Argon} & high & 94.1\% ($\pm$2.1) & 96.4\% ($\pm$1.6) & 91.5\% ($\pm$2.8) & 100.0\% (97.3, 100.0) & 76.6\% ($\pm$6.8) & 23.4 pp ($\pm$6.8) & 1,838 \\
    7 & \textbf{Gemini 4 Argon} & low & 88.7\% ($\pm$2.9) & 93.9\% ($\pm$2.3) & 81.3\% ($\pm$4.1) & 99.8\% (96.8, 100.0) & 56.7\% ($\pm$7.9) & 43.0 pp ($\pm$7.8) & 0 \\
    8 & \textbf{GPT-6.1 Terra} & xhigh & 94.7\% ($\pm$2.1) & 96.6\% ($\pm$1.9) & 91.8\% ($\pm$2.7) & 100.0\% (97.3, 100.0) & 76.4\% ($\pm$6.9) & 23.6 pp ($\pm$6.9) & 622 \\
    8 & \textbf{GPT-6.1 Terra} & low & 91.3\% ($\pm$2.5) & 95.2\% ($\pm$2.1) & 86.3\% ($\pm$3.3) & 100.0\% (97.3, 100.0) & 62.1\% ($\pm$7.7) & 37.9 pp ($\pm$7.7) & 89 \\
    9 & \textbf{Claude Fable 5.2} & max & 93.4\% ($\pm$2.2) & 95.3\% ($\pm$1.9) & 91.5\% ($\pm$2.6) & 100.0\% (97.3, 100.0) & 73.7\% ($\pm$7.3) & 26.3 pp ($\pm$7.3) & 2,951 \\
    9 & \textbf{Claude Fable 5.2} & low & 91.0\% ($\pm$2.5) & 93.7\% ($\pm$2.2) & 87.7\% ($\pm$3.1) & 100.0\% (97.3, 100.0) & 65.5\% ($\pm$7.7) & 34.5 pp ($\pm$7.7) & 125 \\
    10 & \textbf{Gemini 4 Hydrogen} & xhigh & 92.5\% ($\pm$2.4) & 95.6\% ($\pm$1.7) & 88.4\% ($\pm$3.2) & 100.0\% (97.3, 100.0) & 68.9\% ($\pm$7.5) & 31.1 pp ($\pm$7.5) & 13,775 \\
    10 & \textbf{Gemini 4 Hydrogen} & low & 92.9\% ($\pm$2.3) & 95.6\% ($\pm$2.0) & 89.6\% ($\pm$3.0) & 100.0\% (97.3, 100.0) & 70.2\% ($\pm$7.4) & 29.8 pp ($\pm$7.4) & 601 \\
    11 & \textbf{Claude Sonnet 5.5} & max & 91.7\% ($\pm$2.4) & 94.5\% ($\pm$1.9) & 89.2\% ($\pm$2.9) & 100.0\% (97.3, 100.0) & 67.6\% ($\pm$7.7) & 32.4 pp ($\pm$7.7) & 6,340 \\
    11 & \textbf{Claude Sonnet 5.5} & low & 79.3\% ($\pm$3.9) & 86.4\% ($\pm$3.4) & 71.4\% ($\pm$4.7) & 99.3\% (95.9, 99.9) & 35.3\% ($\pm$7.6) & 63.9 pp ($\pm$7.6) & 32 \\
    12 & \textbf{Claude Opus 5.5} & max & 91.2\% ($\pm$2.6) & 94.2\% ($\pm$2.3) & 87.8\% ($\pm$3.0) & 99.3\% (96.0, 99.9) & 63.0\% ($\pm$7.6) & 36.2 pp ($\pm$7.5) & 5,162 \\
    12 & \textbf{Claude Opus 5.5} & low & 89.3\% ($\pm$3.0) & 92.5\% ($\pm$2.5) & 85.1\% ($\pm$3.6) & 99.8\% (96.8, 100.0) & 61.1\% ($\pm$7.9) & 38.7 pp ($\pm$7.9) & 252 \\
    13 & \textbf{Gemini 3.8 Flash} & high & 87.1\% ($\pm$3.2) & 91.9\% ($\pm$2.6) & 81.7\% ($\pm$4.0) & 100.0\% (97.3, 100.0) & 56.8\% ($\pm$8.1) & 43.2 pp ($\pm$8.1) & 8,150 \\
    13 & \textbf{Gemini 3.8 Flash} & low & 82.2\% ($\pm$3.6) & 90.0\% ($\pm$3.0) & 73.4\% ($\pm$4.5) & 100.0\% (97.2, 100.0) & 39.6\% ($\pm$7.7) & 60.4 pp ($\pm$7.7) & 861 \\
    14 & \textbf{Claude Fable 5.1} & max & 85.7\% ($\pm$3.1) & 89.9\% ($\pm$2.7) & 80.7\% ($\pm$3.8) & 100.0\% (97.3, 100.0) & 49.7\% ($\pm$8.0) & 50.3 pp ($\pm$8.0) & 4,665 \\
    14 & \textbf{Claude Fable 5.1} & low & 78.2\% ($\pm$3.9) & 86.5\% ($\pm$3.1) & 69.6\% ($\pm$4.7) & 100.0\% (97.3, 100.0) & 32.4\% ($\pm$7.4) & 67.6 pp ($\pm$7.4) & 120 \\
    15 & \textbf{Gemini 3.6 Flash} & high & 83.8\% ($\pm$3.4) & 90.9\% ($\pm$2.7) & 75.8\% ($\pm$4.4) & 100.0\% (97.3, 100.0) & 44.7\% ($\pm$8.0) & 55.3 pp ($\pm$8.0) & 3,473 \\
    15 & \textbf{Gemini 3.6 Flash} & low & 81.3\% ($\pm$3.5) & 90.3\% ($\pm$2.6) & 70.5\% ($\pm$4.7) & 100.0\% (97.3, 100.0) & 36.6\% ($\pm$7.5) & 63.4 pp ($\pm$7.5) & 1,000 \\
    15 & \textbf{Gemini 3.6 Flash} & minimal & 55.0\% ($\pm$4.8) & 71.2\% ($\pm$4.8) & 39.6\% ($\pm$5.0) & 93.7\% (88.3, 96.7) & 8.0\% ($\pm$4.0) & 85.8 pp ($\pm$5.4) & 0 \\
    16 & \textbf{Claude Haiku 5.5} & max & 83.6\% ($\pm$3.3) & 89.4\% ($\pm$2.8) & 76.3\% ($\pm$4.0) & 99.8\% (96.8, 100.0) & 41.6\% ($\pm$7.9) & 58.1 pp ($\pm$7.9) & 13,351 \\
    16 & \textbf{Claude Haiku 5.5} & low & 68.7\% ($\pm$3.8) & 81.2\% ($\pm$3.5) & 53.9\% ($\pm$4.4) & 99.0\% (95.5, 99.8) & 10.9\% ($\pm$4.5) & 88.1 pp ($\pm$4.7) & 752 \\
    17 & \textbf{GPT-6 Luna} & xhigh & 74.8\% ($\pm$4.1) & 85.4\% ($\pm$3.7) & 60.9\% ($\pm$5.0) & 98.5\% (94.8, 99.6) & 25.6\% ($\pm$6.7) & 73.0 pp ($\pm$6.9) & 1,545 \\
    17 & \textbf{GPT-6 Luna} & low & 63.0\% ($\pm$4.2) & 79.0\% ($\pm$4.1) & 43.3\% ($\pm$4.7) & 98.6\% (94.9, 99.6) & 8.4\% ($\pm$4.1) & 90.2 pp ($\pm$4.3) & 307 \\
    18 & \textbf{Gemini 3.5 Flash-Lite} & high & 68.7\% ($\pm$4.4) & 83.6\% ($\pm$3.6) & 52.6\% ($\pm$5.3) & 98.8\% (95.1, 99.7) & 18.6\% ($\pm$6.0) & 80.1 pp ($\pm$6.1) & 3,094 \\
    18 & \textbf{Gemini 3.5 Flash-Lite} & low & 58.5\% ($\pm$4.7) & 74.4\% ($\pm$4.6) & 41.4\% ($\pm$5.0) & 95.0\% (90.0, 97.6) & 8.6\% ($\pm$4.4) & 86.4 pp ($\pm$5.3) & 840 \\
    18 & \textbf{Gemini 3.5 Flash-Lite} & minimal & 37.9\% ($\pm$4.1) & 56.9\% ($\pm$4.8) & 20.8\% ($\pm$3.8) & 89.2\% (82.9, 93.4) & 0.9\% ($\pm$1.2) & 88.3 pp ($\pm$5.1) & 0 \\
    \midrule
    19* & \textbf{GPT-5.6 Sol} & max & 87.2\% ($\pm$3.4) & 91.5\% ($\pm$3.0) & 81.9\% ($\pm$4.0) & 98.3\% (94.4, 99.5) & 54.2\% ($\pm$7.9) & 44.0 pp ($\pm$8.0) & 1,653 \\
    20* & \textbf{GPT-5.5} & xhigh & 84.5\% ($\pm$3.6) & 89.4\% ($\pm$2.9) & 79.2\% ($\pm$4.3) & 99.9\% (97.0, 100.0) & 51.7\% ($\pm$8.0) & 48.2 pp ($\pm$8.0) & 2,299 \\
    21* & \textbf{Claude Opus 5} & max & 81.3\% ($\pm$4.0) & 85.8\% ($\pm$3.6) & 76.8\% ($\pm$4.5) & 99.8\% (96.8, 100.0) & 47.3\% ($\pm$8.1) & 52.4 pp ($\pm$8.1) & 2,326 \\
    22* & \textbf{Gemini 3.1 Pro} & high & 83.6\% ($\pm$3.5) & 92.2\% ($\pm$2.5) & 73.1\% ($\pm$4.8) & 100.0\% (97.3, 100.0) & 43.6\% ($\pm$7.7) & 56.4 pp ($\pm$7.7) & 2,834 \\
    22* & \textbf{Gemini 3.1 Pro} & low & 83.3\% ($\pm$3.6) & 91.6\% ($\pm$2.5) & 73.3\% ($\pm$5.0) & 100.0\% (97.3, 100.0) & 46.1\% ($\pm$8.0) & 53.9 pp ($\pm$8.0) & 1,219 \\
    23* & \textbf{GPT-5.6 Terra} & max & 81.6\% ($\pm$3.7) & 88.8\% ($\pm$3.3) & 72.6\% ($\pm$4.6) & 99.2\% (95.9, 99.9) & 39.0\% ($\pm$7.7) & 60.2 pp ($\pm$7.7) & 3,501 \\
    24* & \textbf{Claude Opus 4.8} & max & 71.3\% ($\pm$5.0) & 80.0\% ($\pm$4.2) & 62.0\% ($\pm$5.9) & 99.1\% (95.7, 99.8) & 36.9\% ($\pm$7.9) & 62.2 pp ($\pm$7.9) & 4,613 \\
    25* & \textbf{Claude Sonnet 5} & max & 66.6\% ($\pm$5.5) & 76.2\% ($\pm$4.8) & 57.1\% ($\pm$6.3) & 97.7\% (93.6, 99.2) & 32.9\% ($\pm$7.8) & 64.8 pp ($\pm$7.7) & 12,225 \\
    26* & \textbf{Gemini 3 Flash} & high & 77.3\% ($\pm$3.9) & 88.5\% ($\pm$3.0) & 64.9\% ($\pm$5.0) & 100.0\% (97.3, 100.0) & 31.7\% ($\pm$7.3) & 68.3 pp ($\pm$7.3) & 4,071 \\
    26* & \textbf{Gemini 3 Flash} & low & 74.5\% ($\pm$4.1) & 86.7\% ($\pm$3.3) & 60.7\% ($\pm$5.2) & 99.0\% (95.6, 99.8) & 26.0\% ($\pm$6.9) & 73.1 pp ($\pm$7.0) & 1,016 \\
    27* & \textbf{GPT-5.6 Luna} & max & 75.9\% ($\pm$4.0) & 85.4\% ($\pm$3.5) & 64.4\% ($\pm$4.9) & 98.8\% (95.2, 99.7) & 28.6\% ($\pm$7.2) & 70.2 pp ($\pm$7.2) & 4,198 \\
    28* & \textbf{Claude Sonnet 4.6} & max & 60.0\% ($\pm$5.7) & 68.8\% ($\pm$5.4) & 50.6\% ($\pm$6.2) & 92.9\% (87.3, 96.1) & 24.6\% ($\pm$7.0) & 68.3 pp ($\pm$7.5) & 6,240 \\
    29* & \textbf{Gemini 3.1 Flash-Lite} & high & 64.7\% ($\pm$4.8) & 79.0\% ($\pm$4.3) & 49.2\% ($\pm$5.5) & 96.9\% (92.5, 98.8) & 18.3\% ($\pm$6.1) & 78.6 pp ($\pm$6.5) & 3,381 \\
    29* & \textbf{Gemini 3.1 Flash-Lite} & low & 39.0\% ($\pm$3.8) & 59.4\% ($\pm$4.7) & 20.9\% ($\pm$3.6) & 92.2\% (86.4, 95.6) & 0.0\% ($\pm$0.0) & 92.2 pp ($\pm$4.1) & 126 \\
    \bottomrule
  \end{tabular}%
  }
  \tablenotes{Thought-token counts reflect each provider's native tokenizer and API reporting (e.g., OpenAI \texttt{reasoning\_tokens}; Gemini~4 Argon low reports 0 thought tokens) and are not directly comparable across providers. Models are ordered by descending family pass\hfecaret{}4 at highest evaluated reasoning effort (current tier first, then prior-generation models marked with *). Reported $\pm$ values denote half-widths $(\mathrm{hi} - \mathrm{lo})/2$ of the 95\% family-clustered percentile bootstrap interval. *~Prior-generation models, reported for reference.}
\end{table}

\clearpage

\begin{table}[!htbp]
  \centering
  \caption{Effect of reasoning effort on reliability. Each model is shown at its highest evaluated effort (\mbox{high/max/xhigh}), at low effort, and, where available, at minimal effort (0 thought tokens).}
  \label{tab:effort}
  \resizebox{\textwidth}{!}{%
  \renewcommand{\arraystretch}{0.96}
  \begin{tabular}{@{}l l c c c c c c c r c c c c@{}}
    \toprule
    \textbf{Model} & \textbf{Effort} & \thead{\textbf{pass@1}\\\textbf{($\pm$CI)}} & \thead{\textbf{pass@4}\\\textbf{($\pm$CI)}} & \thead{\textbf{pass\hfecaret{}4 / Rel@4}\\\textbf{($\pm$CI)}} & \thead{\textbf{family}\\\textbf{pass@4}} & \thead{\textbf{family pass\hfecaret{}4}\\\textbf{($\pm$CI)}} & \thead{\textbf{$\Delta$ family pass\hfecaret{}4,}\\\textbf{this effort $-$ low}} & \thead{\textbf{$\Delta$ family pass\hfecaret{}4,}\\\textbf{this effort $-$ minimal}} & \thead{\textbf{Avg Thought}\\\textbf{Tokens}} & \thead{\textbf{Math fam pass\hfecaret{}4}\\\textbf{($N=160$)}} & \thead{\textbf{Reas fam pass\hfecaret{}4}\\\textbf{($N=72$)}} & \thead{\textbf{Spatial fam pass\hfecaret{}4}\\\textbf{($N=108$)}} & \thead{\textbf{Vision fam pass\hfecaret{}4}\\\textbf{($N=204$)}} \\
    \midrule
    \textbf{GPT-6.2 Astra} & xhigh & 97.8\% ($\pm$1.5) & 97.9\% ($\pm$1.5) & 97.5\% ($\pm$1.6) & 100.0\% & \textbf{92.4\%} ($\pm$4.3) & \textbf{+18.1 pp} & --- & 204 & 98.3\% & 88.9\% & 90.1\% & 90.2\% \\
    \textbf{GPT-6.2 Astra} & low & 94.9\% ($\pm$1.9) & 97.9\% ($\pm$1.5) & 90.7\% ($\pm$2.9) & 100.0\% & 74.3\% ($\pm$7.0) & --- & --- & 14 & 57.2\% & 88.9\% & 72.3\% & 83.7\% \\
    \addlinespace[1.5pt]
    \textbf{GPT-6.2 Sol} & xhigh & 97.3\% ($\pm$1.5) & 98.4\% ($\pm$1.2) & 95.8\% ($\pm$1.9) & 100.0\% & \textbf{86.9\%} ($\pm$5.2) & \textbf{+18.2 pp} & --- & 274 & 96.0\% & 83.7\% & 85.6\% & 81.7\% \\
    \textbf{GPT-6.2 Sol} & low & 93.1\% ($\pm$2.5) & 96.8\% ($\pm$1.9) & 87.3\% ($\pm$3.7) & 100.0\% & 68.7\% ($\pm$7.1) & --- & --- & 15 & 57.2\% & 59.6\% & 62.3\% & 84.3\% \\
    \addlinespace[1.5pt]
    \textbf{Gemini 4 Neon} & xhigh & 96.8\% ($\pm$1.5) & 97.9\% ($\pm$1.3) & 95.6\% ($\pm$1.9) & 100.0\% & \textbf{85.8\%} ($\pm$5.8) & \textbf{$-$1.0 pp} & --- & 9,852 & 97.5\% & 83.3\% & 86.4\% & 77.3\% \\
    \textbf{Gemini 4 Neon} & low & 96.9\% ($\pm$1.7) & 97.6\% ($\pm$1.4) & 96.0\% ($\pm$1.9) & 100.0\% & 86.8\% ($\pm$5.5) & --- & --- & 787 & 95.8\% & 88.9\% & 81.5\% & 81.8\% \\
    \addlinespace[1.5pt]
    \textbf{GPT-6 Astra} & xhigh & 96.8\% ($\pm$1.6) & 98.0\% ($\pm$1.4) & 94.9\% ($\pm$2.0) & 100.0\% & \textbf{83.4\%} ($\pm$5.8) & \textbf{+2.5 pp} & --- & 382 & 92.5\% & 88.9\% & 84.0\% & 73.9\% \\
    \textbf{GPT-6 Astra} & low & 96.1\% ($\pm$1.8) & 97.9\% ($\pm$1.5) & 93.8\% ($\pm$2.3) & 100.0\% & 80.9\% ($\pm$6.3) & --- & --- & 65 & 91.7\% & 85.6\% & 72.2\% & 75.4\% \\
    \addlinespace[1.5pt]
    \textbf{Gemini 4 Carbon} & high & 95.6\% ($\pm$1.7) & 97.0\% ($\pm$1.3) & 93.8\% ($\pm$2.2) & 100.0\% & \textbf{80.5\%} ($\pm$6.4) & \textbf{+17.4 pp} & --- & 6,616 & 97.5\% & 83.7\% & 73.1\% & 69.9\% \\
    \textbf{Gemini 4 Carbon} & low & 90.9\% ($\pm$2.5) & 95.6\% ($\pm$1.8) & 84.4\% ($\pm$3.9) & 100.0\% & 63.1\% ($\pm$7.7) & --- & --- & 29 & 55.1\% & 63.1\% & 67.9\% & 66.8\% \\
    \addlinespace[1.5pt]
    \textbf{GPT-6.1 Sol} & xhigh & 96.1\% ($\pm$1.7) & 97.8\% ($\pm$1.6) & 93.7\% ($\pm$2.2) & 100.0\% & \textbf{79.1\%} ($\pm$6.4) & \textbf{+11.6 pp} & --- & 396 & 93.3\% & 80.4\% & 78.4\% & 67.9\% \\
    \textbf{GPT-6.1 Sol} & low & 94.0\% ($\pm$2.0) & 96.7\% ($\pm$1.7) & 90.0\% ($\pm$2.5) & 100.0\% & 67.5\% ($\pm$7.3) & --- & --- & 79 & 84.7\% & 67.0\% & 57.0\% & 59.7\% \\
    \addlinespace[1.5pt]
    \textbf{Gemini 4 Argon} & high & 94.1\% ($\pm$2.1) & 96.4\% ($\pm$1.6) & 91.5\% ($\pm$2.8) & 100.0\% & \textbf{76.6\%} ($\pm$6.8) & \textbf{+19.9 pp} & --- & 1,838 & 87.0\% & 88.9\% & 71.6\% & 66.7\% \\
    \textbf{Gemini 4 Argon} & low & 88.7\% ($\pm$2.9) & 93.9\% ($\pm$2.3) & 81.3\% ($\pm$4.1) & 99.8\% & 56.7\% ($\pm$7.9) & --- & --- & 0 & 47.8\% & 52.8\% & 53.3\% & 66.9\% \\
    \addlinespace[1.5pt]
    \textbf{GPT-6.1 Terra} & xhigh & 94.7\% ($\pm$2.1) & 96.6\% ($\pm$1.9) & 91.8\% ($\pm$2.7) & 100.0\% & \textbf{76.4\%} ($\pm$6.9) & \textbf{+14.2 pp} & --- & 622 & 86.6\% & 74.2\% & 72.0\% & 71.4\% \\
    \textbf{GPT-6.1 Terra} & low & 91.3\% ($\pm$2.5) & 95.2\% ($\pm$2.1) & 86.3\% ($\pm$3.3) & 100.0\% & 62.1\% ($\pm$7.7) & --- & --- & 89 & 53.9\% & 66.8\% & 58.4\% & 68.9\% \\
    \addlinespace[1.5pt]
    \textbf{Claude Fable 5.2} & max & 93.4\% ($\pm$2.2) & 95.3\% ($\pm$1.9) & 91.5\% ($\pm$2.6) & 100.0\% & \textbf{73.7\%} ($\pm$7.3) & \textbf{+8.2 pp} & --- & 2,951 & 83.3\% & 83.3\% & 74.3\% & 62.4\% \\
    \textbf{Claude Fable 5.2} & low & 91.0\% ($\pm$2.5) & 93.7\% ($\pm$2.2) & 87.7\% ($\pm$3.1) & 100.0\% & 65.5\% ($\pm$7.7) & --- & --- & 125 & 77.7\% & 76.3\% & 53.8\% & 58.3\% \\
    \addlinespace[1.5pt]
    \textbf{Gemini 4 Hydrogen} & xhigh & 92.5\% ($\pm$2.4) & 95.6\% ($\pm$1.7) & 88.4\% ($\pm$3.2) & 100.0\% & \textbf{68.9\%} ($\pm$7.5) & \textbf{$-$1.3 pp} & --- & 13,775 & 93.3\% & 83.7\% & 62.4\% & 47.9\% \\
    \textbf{Gemini 4 Hydrogen} & low & 92.9\% ($\pm$2.3) & 95.6\% ($\pm$2.0) & 89.6\% ($\pm$3.0) & 100.0\% & 70.2\% ($\pm$7.4) & --- & --- & 601 & 85.5\% & 88.9\% & 50.4\% & 62.1\% \\
    \addlinespace[1.5pt]
    \textbf{Claude Sonnet 5.5} & max & 91.7\% ($\pm$2.4) & 94.5\% ($\pm$1.9) & 89.2\% ($\pm$2.9) & 100.0\% & \textbf{67.6\%} ($\pm$7.7) & \textbf{+32.3 pp} & --- & 6,340 & 85.3\% & 78.1\% & 56.8\% & 55.7\% \\
    \textbf{Claude Sonnet 5.5} & low & 79.3\% ($\pm$3.9) & 86.4\% ($\pm$3.4) & 71.4\% ($\pm$4.7) & 99.3\% & 35.3\% ($\pm$7.6) & --- & --- & 32 & 43.9\% & 34.5\% & 22.1\% & 35.9\% \\
    \addlinespace[1.5pt]
    \textbf{Claude Opus 5.5} & max & 91.2\% ($\pm$2.6) & 94.2\% ($\pm$2.3) & 87.8\% ($\pm$3.0) & 99.3\% & \textbf{63.0\%} ($\pm$7.6) & \textbf{+2.0 pp} & --- & 5,162 & 75.2\% & 79.6\% & 55.8\% & 51.5\% \\
    \textbf{Claude Opus 5.5} & low & 89.3\% ($\pm$3.0) & 92.5\% ($\pm$2.5) & 85.1\% ($\pm$3.6) & 99.8\% & 61.1\% ($\pm$7.9) & --- & --- & 252 & 72.8\% & 76.3\% & 50.8\% & 51.9\% \\
    \addlinespace[1.5pt]
    \textbf{Gemini 3.8 Flash} & high & 87.1\% ($\pm$3.2) & 91.9\% ($\pm$2.6) & 81.7\% ($\pm$4.0) & 100.0\% & \textbf{56.8\%} ($\pm$8.1) & \textbf{+17.2 pp} & --- & 8,150 & 76.8\% & 69.6\% & 46.4\% & 42.0\% \\
    \textbf{Gemini 3.8 Flash} & low & 82.2\% ($\pm$3.6) & 90.0\% ($\pm$3.0) & 73.4\% ($\pm$4.5) & 100.0\% & 39.6\% ($\pm$7.7) & --- & --- & 861 & 44.7\% & 30.5\% & 31.6\% & 43.0\% \\
    \addlinespace[1.5pt]
    \textbf{Claude Fable 5.1} & max & 85.7\% ($\pm$3.1) & 89.9\% ($\pm$2.7) & 80.7\% ($\pm$3.8) & 100.0\% & \textbf{49.7\%} ($\pm$8.0) & \textbf{+17.2 pp} & --- & 4,665 & 70.8\% & 81.5\% & 33.4\% & 30.5\% \\
    \textbf{Claude Fable 5.1} & low & 78.2\% ($\pm$3.9) & 86.5\% ($\pm$3.1) & 69.6\% ($\pm$4.7) & 100.0\% & 32.4\% ($\pm$7.4) & --- & --- & 120 & 45.8\% & 59.3\% & 18.8\% & 19.7\% \\
    \addlinespace[1.5pt]
    \textbf{Gemini 3.6 Flash} & high & 83.8\% ($\pm$3.4) & 90.9\% ($\pm$2.7) & 75.8\% ($\pm$4.4) & 100.0\% & \textbf{44.7\%} ($\pm$8.0) & \textbf{+8.1 pp} & \textbf{+36.8 pp} & 3,473 & 69.6\% & 48.5\% & 34.6\% & 29.2\% \\
    \textbf{Gemini 3.6 Flash} & low & 81.3\% ($\pm$3.5) & 90.3\% ($\pm$2.6) & 70.5\% ($\pm$4.7) & 100.0\% & 36.6\% ($\pm$7.5) & --- & +28.6 pp & 1,000 & 56.5\% & 28.8\% & 24.0\% & 30.5\% \\
    \textbf{Gemini 3.6 Flash} & minimal & 55.0\% ($\pm$4.8) & 71.2\% ($\pm$4.8) & 39.6\% ($\pm$5.0) & 93.7\% & 8.0\% ($\pm$4.0) & --- & --- & 0 & 2.2\% & 5.9\% & 4.8\% & 14.9\% \\
    \addlinespace[1.5pt]
    \textbf{Gemini 3.1 Pro*} & high & 83.6\% ($\pm$3.5) & 92.2\% ($\pm$2.5) & 73.1\% ($\pm$4.8) & 100.0\% & \textbf{43.6\%} ($\pm$7.7) & \textbf{$-$2.5 pp} & --- & 2,834 & 80.9\% & 43.3\% & 31.1\% & 21.1\% \\
    \textbf{Gemini 3.1 Pro*} & low & 83.3\% ($\pm$3.6) & 91.6\% ($\pm$2.5) & 73.3\% ($\pm$5.0) & 100.0\% & 46.1\% ($\pm$8.0) & --- & --- & 1,219 & 73.7\% & 42.0\% & 42.1\% & 28.1\% \\
    \addlinespace[1.5pt]
    \textbf{Claude Haiku 5.5} & max & 83.6\% ($\pm$3.3) & 89.4\% ($\pm$2.8) & 76.3\% ($\pm$4.0) & 99.8\% & \textbf{41.6\%} ($\pm$7.9) & \textbf{+30.7 pp} & --- & 13,351 & 70.3\% & 37.8\% & 31.0\% & 26.1\% \\
    \textbf{Claude Haiku 5.5} & low & 68.7\% ($\pm$3.8) & 81.2\% ($\pm$3.5) & 53.9\% ($\pm$4.4) & 99.0\% & 10.9\% ($\pm$4.5) & --- & --- & 752 & 21.1\% & 9.3\% & 7.4\% & 5.4\% \\
    \addlinespace[1.5pt]
    \textbf{Gemini 3 Flash*} & high & 77.3\% ($\pm$3.9) & 88.5\% ($\pm$3.0) & 64.9\% ($\pm$5.0) & 100.0\% & \textbf{31.7\%} ($\pm$7.3) & \textbf{+5.7 pp} & --- & 4,071 & 66.9\% & 17.5\% & 25.2\% & 12.4\% \\
    \textbf{Gemini 3 Flash*} & low & 74.5\% ($\pm$4.1) & 86.7\% ($\pm$3.3) & 60.7\% ($\pm$5.2) & 99.0\% & 26.0\% ($\pm$6.9) & --- & --- & 1,016 & 48.6\% & 13.3\% & 16.5\% & 17.7\% \\
    \addlinespace[1.5pt]
    \textbf{GPT-6 Luna} & xhigh & 74.8\% ($\pm$4.1) & 85.4\% ($\pm$3.7) & 60.9\% ($\pm$5.0) & 98.5\% & \textbf{25.6\%} ($\pm$6.7) & \textbf{+17.2 pp} & --- & 1,545 & 30.0\% & 52.2\% & 9.2\% & 21.3\% \\
    \textbf{GPT-6 Luna} & low & 63.0\% ($\pm$4.2) & 79.0\% ($\pm$4.1) & 43.3\% ($\pm$4.7) & 98.6\% & 8.4\% ($\pm$4.1) & --- & --- & 307 & 13.2\% & 11.5\% & 1.5\% & 7.1\% \\
    \addlinespace[1.5pt]
    \textbf{Gemini 3.5 Flash-Lite} & high & 68.7\% ($\pm$4.4) & 83.6\% ($\pm$3.6) & 52.6\% ($\pm$5.3) & 98.8\% & \textbf{18.6\%} ($\pm$6.0) & \textbf{+10.0 pp} & \textbf{+17.7 pp} & 3,094 & 46.5\% & 17.2\% & 5.7\% & 4.1\% \\
    \textbf{Gemini 3.5 Flash-Lite} & low & 58.5\% ($\pm$4.7) & 74.4\% ($\pm$4.6) & 41.4\% ($\pm$5.0) & 95.0\% & 8.6\% ($\pm$4.4) & --- & +7.7 pp & 840 & 27.8\% & 0.5\% & 0.4\% & 0.8\% \\
    \textbf{Gemini 3.5 Flash-Lite} & minimal & 37.9\% ($\pm$4.1) & 56.9\% ($\pm$4.8) & 20.8\% ($\pm$3.8) & 89.2\% & 0.9\% ($\pm$1.2) & --- & --- & 0 & 0.2\% & 5.6\% & 0.2\% & 0.2\% \\
    \addlinespace[1.5pt]
    \textbf{Gemini 3.1 Flash-Lite*} & high & 64.7\% ($\pm$4.8) & 79.0\% ($\pm$4.3) & 49.2\% ($\pm$5.5) & 96.9\% & \textbf{18.3\%} ($\pm$6.1) & \textbf{+18.3 pp} & --- & 3,381 & 48.6\% & 13.6\% & 2.1\% & 4.7\% \\
    \textbf{Gemini 3.1 Flash-Lite*} & low & 39.0\% ($\pm$3.8) & 59.4\% ($\pm$4.7) & 20.9\% ($\pm$3.6) & 92.2\% & 0.0\% ($\pm$0.0) & --- & --- & 126 & 0.0\% & 0.0\% & 0.0\% & 0.0\% \\
    \bottomrule
  \end{tabular}%
  }
  \tablenotes{The $\Delta$ columns give the change in family pass\hfecaret{}4 of that row's effort level relative to the same model's low or minimal run. *~Prior-generation models, reported for reference. Models are ordered by descending family pass\hfecaret{}4 at highest evaluated reasoning effort. Reported $\pm$ values denote half-widths $(\mathrm{hi} - \mathrm{lo})/2$ of the 95\% family-clustered percentile bootstrap interval.}
\end{table}

\clearpage

\begin{table}[!htbp]
  \centering
  \caption{Subcategory sensitivity to thinking effort (20 subcategories ranked by mean $\Delta$ family pass\hfecaret{}4, \mbox{high/max/xhigh} $-$ low, across 18 models).}
  \label{tab:subcat_effort}
  \footnotesize
  \setlength{\tabcolsep}{4pt}
  \renewcommand{\arraystretch}{1.15}
  \begin{tabular*}{\textwidth}{@{}r@{\hspace{8pt}} l@{\extracolsep{\fill}} l r r@{}}
    \toprule
    \textbf{Rank} & \textbf{Subcategory} & \textbf{Macro-Category} & \textbf{Families} & \begin{tabular}[b]{@{}r@{}}\textbf{Mean $\Delta$ family pass\hfecaret{}4}\\\textbf{(High/Max/xHigh $-$ Low,}\\\textbf{18 Models)}\end{tabular} \\
    \midrule
    1 & \textbf{Logical Deduction} & Reasoning \& Language & 2 & +29.7 pp \\
    2 & \textbf{Constraint Satisfaction} & Math \& Algorithmic & 13 & +28.5 pp \\
    3 & \textbf{Algorithmic and Symbolic Execution} & Math \& Algorithmic & 8 & +26.2 pp \\
    4 & \textbf{Sequence Manipulation} & Math \& Algorithmic & 6 & +26.0 pp \\
    5 & \textbf{Sorting and Ordering} & Math \& Algorithmic & 5 & +21.6 pp \\
    6 & \textbf{General Arithmetic and Calculation} & Math \& Algorithmic & 3 & +19.6 pp \\
    7 & \textbf{Negative Fact Verification} & Reasoning \& Language & 3 & +19.6 pp \\
    8 & \textbf{Positional Tracking} & Spatial \& Positional & 6 & +18.7 pp \\
    9 & \textbf{Causal and Counterfactual Reasoning} & Reasoning \& Language & 2 & +18.3 pp \\
    10 & \textbf{Generalization and Pattern Recognition} & Reasoning \& Language & 4 & +18.3 pp \\
    11 & \textbf{Spatial and Topological Reasoning} & Spatial \& Positional & 4 & +15.8 pp \\
    12 & \textbf{Commonsense and Physical Reasoning} & Reasoning \& Language & 4 & +13.8 pp \\
    13 & \textbf{Spatial Reasoning} & Spatial \& Positional & 17 & +13.1 pp \\
    14 & \textbf{Pragmatic Reasoning} & Reasoning \& Language & 3 & +10.7 pp \\
    15 & \textbf{Text \& Symbol Recognition} & Vision & 11 & +5.7 pp \\
    16 & \textbf{Graph \& Chart Understanding} & Vision & 7 & +5.0 pp \\
    17 & \textbf{Mathematical and Quantitative Reasoning} & Math \& Algorithmic & 5 & +4.9 pp \\
    18 & \textbf{Color Identification} & Vision & 10 & +4.9 pp \\
    19 & \textbf{Arrows, pointers, or indicators} & Vision & 11 & +2.0 pp \\
    20 & \textbf{Object Counting} & Vision & 12 & +1.9 pp \\
    \bottomrule
  \end{tabular*}
\end{table}

\clearpage
\subsection*{Example Four-Prompt Task Family (Public Split)}

\begin{figure}[H]
  \centering
  \includegraphics[width=0.78\textwidth, trim=0 2pt 0 0, clip]{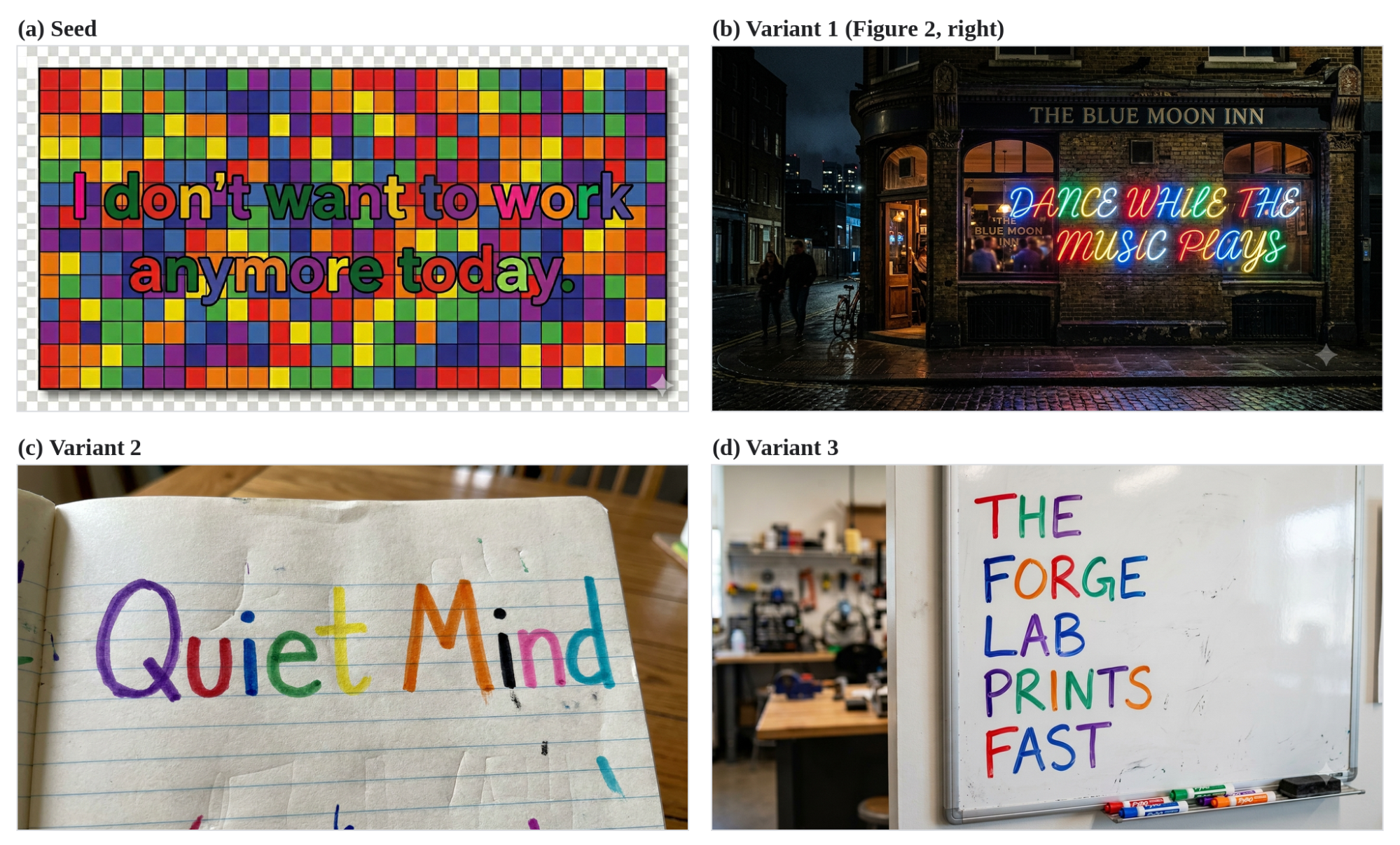}
  \caption{All four images for the Color Identification task family (public split). Table~\ref{tab:color_family} lists the corresponding prompts and ground-truth answers.}
  \label{fig:color_family}
\end{figure}

To illustrate how HFE pairs each seed prompt with three logic-preserving variants, Figure~\ref{fig:color_family} shows the complete four-prompt Color Identification family highlighted in Figure~\ref{fig:examples} (right), whose seed was solved by 4/4 human raters.

\begin{table}[H]
  \centering
  \caption{Prompts and ground-truth answers for the four-prompt Color Identification family in Figure~\ref{fig:color_family}.}
  \label{tab:color_family}
  \footnotesize
  \setlength{\tabcolsep}{5pt}
  \renewcommand{\arraystretch}{1.15}
  \begin{tabularx}{\textwidth}{@{}l >{\raggedright\arraybackslash}X >{\raggedright\arraybackslash}p{0.28\textwidth}@{}}
    \toprule
    \textbf{Panel} & \textbf{Prompt} & \textbf{Ground Truth} \\
    \midrule
    \textbf{(a) Seed} & What color is the second letter in every word? If the word only has one letter, ignore that word. Return your answer by listing only the identified colors from the beginning of the sentence to the end. & Red; Green; Purple; Orange; Green; Red \\
    \textbf{(b) Variant 1 (Fig.~\ref{fig:examples}, right)} & List the color of the third letter of each word in the neon sign in order, separated by a semicolon. & green; yellow; white; blue; blue \\
    \textbf{(c) Variant 2} & List the colors of every vowel in the phrase, separated by commas. & red, blue, green, black \\
    \textbf{(d) Variant 3} & List the color of the second consonant of every word on the whiteboard, in order from top to bottom, separated by a comma. & green, red, blue, green, blue \\
    \bottomrule
  \end{tabularx}
\end{table}

\clearpage
\subsection*{Autorater Grading Prompt}

Below is the verbatim grading prompt template used by the deterministic Gemini~3.1 Pro autorater (\S\ref{sec:grading}), evaluated after stripping internal reasoning traces and control tokens:

\begin{lstlisting}[
  basicstyle=\ttfamily\footnotesize,
  upquote=true,
  breaklines=true,
  breakatwhitespace=true,
  breakindent=0pt,
  columns=fullflexible,
  keepspaces=true,
  frame=single,
  xleftmargin=4pt,
  xrightmargin=4pt
]
[Task]: You are an expert tasked with grading a puzzle solution.
You will be provided with:
1. A question that was used as a prompt for the model.
2. A model's response
3. A reference answer (ground truth).

Your role is to determine whether the final answer in the model's response is equivalent to the reference answer. Respond only with Yes or No.

[Guidelines for Equivalence]:
1) Answer Choices and Values: If the reference answer is a choice (e.g., B), and model's response is enclosed in brackets (e.g., (B), [B], or $\boxed[B]$, consider the answer to be correct (Yes).
2) Numerics: If the reference answer is a numeric value (e.g., 10), and the model's response is the same numeric value spelled out in words (e.g., ten), or vice-versa, consider the answer to be correct (Yes).
3) Temporal quantities: If the reference answer is a temporal quantity at a certain resolution (e.g., 3 hours), and the model's response is at a different resolution (e.g., 180 minutes) but the two are equivalent (e.g., 3 hours is 180 minutes), consider the answer to be correct (Yes).
4) Rubrics: If the ground truth is a rubric (e.g., the response must contain exactly 10 words) and the model's response satisfies all the criteria in the rubric, consider the answer to be correct (Yes).

[Input]:
Question:
{prompt}
-------------------
Model response:
{response}
-------------------
Ground truth:
{target}

[Output]:
Is the final answer in the model's response equivalent to the reference answer, answer Yes or No:
\end{lstlisting}

\clearpage
\subsection*{Full 18-Model Family Reliability Gap}

\begin{figure}[H]
  \centering
  \includegraphics[width=0.96\textwidth]{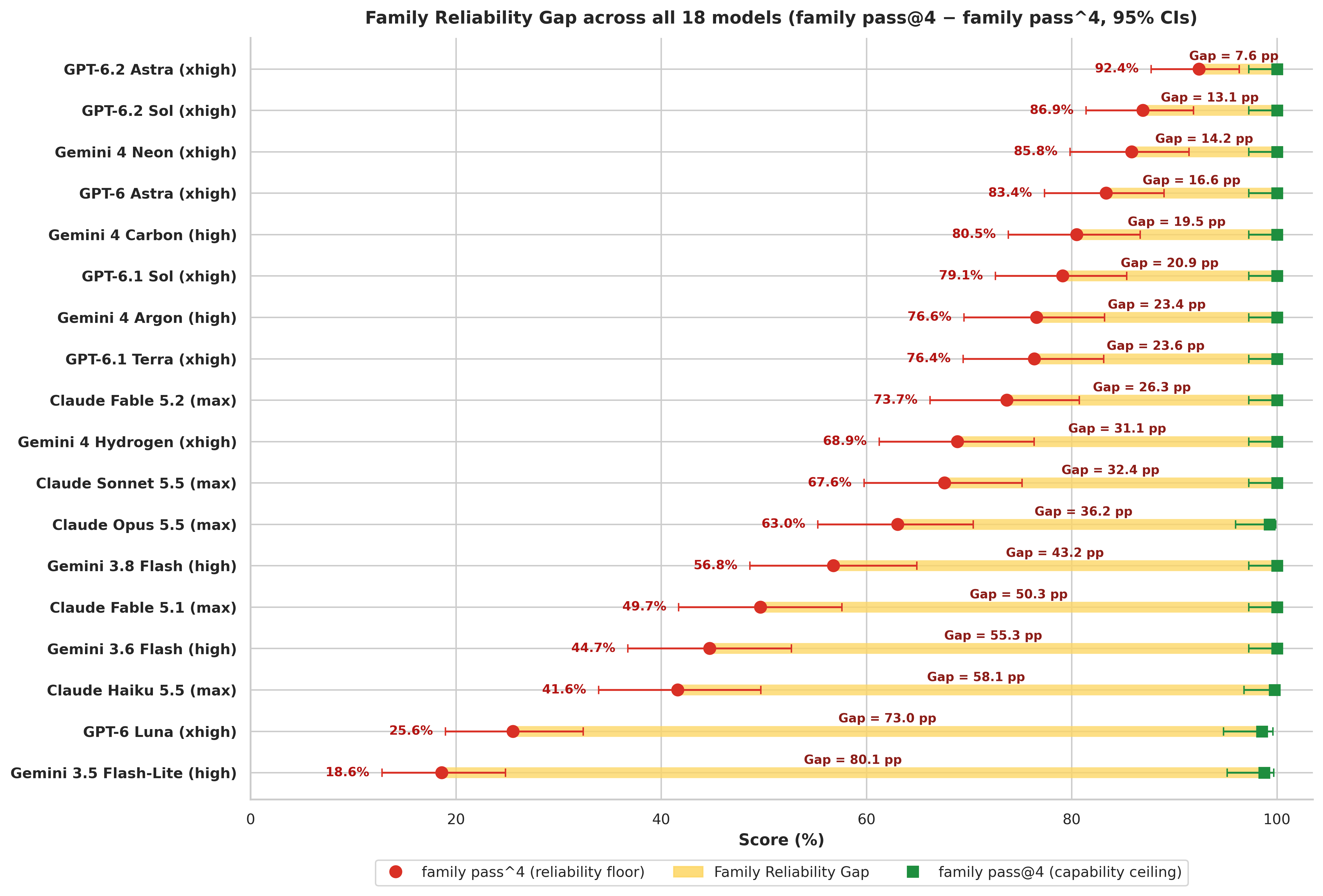}
  \caption{Family Reliability Gap (family pass@4 $-$ family pass\hfecaret{}4, with 95\% CIs) across all 18 models in Table~\ref{tab:main_results} at highest evaluated reasoning effort.}
  \label{fig:reliability_gap_full18}
\end{figure}

\end{document}